\documentclass[]{interact}

\usepackage{epstopdf}
\usepackage[caption=false]{subfig}
\usepackage[numbers,sort&compress]{natbib}
\bibpunct[, ]{[}{]}{,}{n}{,}{,}
\renewcommand\bibfont{\fontsize{10}{12}\selectfont}
\makeatletter
\def\NAT@def@citea{\def\@citea{\NAT@separator}}
\makeatother

\theoremstyle{plain}

\theoremstyle{definition}

\theoremstyle{remark}

\usepackage{amsmath,amsfonts}
\usepackage{algorithmic}
\usepackage{algorithm}
\usepackage{array}
\usepackage{textcomp}
\usepackage{stfloats}
\usepackage{url}
\usepackage{verbatim}
\usepackage{graphicx}
\usepackage{amssymb}
\usepackage{mathtools}
\usepackage{enumitem}
\usepackage{dashbox}
\usepackage{multirow}
\usepackage[table]{xcolor}

\renewtagform{default}{\normalsize(}{)}
\usetagform{default}

\begin{document}

\title{Connectivity Preserving Decentralized UAV Swarm Navigation in Obstacle-laden Environments without Explicit Communication}

\author{
    \name{Thiviyathinesvaran Palani$^1$, Hiroaki Fukushima$^1$\thanks{CONTACT Hiroaki Fukushima. Email: fukushima.hiroaki@kuas.ac.jp}, and Shunsuke Izuhara$^2$}
    \affil{$^1$Graduate School of Engineering, Kyoto University of Advanced Science, Kyoto, Japan; %
           $^2$Faculty of Environmental, Life, Natural Science and Technology, Okayama University, Okayama, Japan}
}

\maketitle

\begin{abstract}
    This paper presents a novel control method for a group of UAVs in obstacle-laden environments while preserving sensing network connectivity without data transmission between the UAVs.
    By leveraging constraints rooted in control barrier functions (CBFs), the proposed method aims to overcome the limitations, such as oscillatory behaviors and frequent constraint violations, of the existing method based on artificial potential fields (APFs).
    More specifically, the proposed method first determines desired control inputs by considering CBF-based constraints rather than repulsive APFs.
    The desired inputs are then minimally modified by solving a numerical optimization problem with soft constraints.
    In addition to the optimization-based method, we present an approximate method without numerical optimization.
    The effectiveness of the proposed methods is evaluated by extensive simulations to compare the performance of the CBF-based methods with an APF-based approach.
    Experimental results using real quadrotors are also presented.
\end{abstract}

\begin{keywords}
    Leader-follower navigation,
    Non-communicative,
    Control barrier function,
    Connectivity preservation,
    Collision avoidance
\end{keywords}

\section{Introduction} \label{sec:intro}
The realm of robotics has witnessed significant advancements through research on cooperative control of multiple robot systems, driving technological advancements for a wide range of applications \cite{Parker2016, Mesbahi2010, Dorigo2021}.
This paper focuses on the fundamental problem of how to move a group of robots as a whole to a target area.
Specifically, we assume that only one of the robots, called the leader, knows the path to the target area.
Thus, since each robot in the group has a limited sensing and communication range, the connectivity of the sensing/communication network must be preserved to avoid leaving some robots behind.
Furthermore, to move through cluttered environments including narrow spaces without getting stuck, robots need to flexibly change their network topology in a decentralized way while preserving the network connectivity.

To overcome this problem and to deal with environments where no wireless network is available for information exchange, Sakai et al. \cite{Sakai2018} proposed a method that does not rely on data transmission between robots.
This method can change network topology by deactivating certain network links to allow the robots to pass through narrow spaces while enhancing active links in open areas to maintain group cohesion.
The method proposed in \cite{Nomura2021} extended the approach from \cite{Sakai2018} for 2D environments to cater to multiple UAVs in 3D environments.
However, the method proposed in \cite{Nomura2021} often suffers from oscillatory movements and frequent constraint violations, which could be a serious problem especially for UAVs.
To overcome these problems, we aim to develop a new algorithm inspired by control barrier functions (CBFs), which have extensively studied in recent years.

\subsection{Related work}
CBFs were initially introduced to deal with inequality constraints for collision avoidance \cite{Ames2017, WangL2017, Chen2021, Ames2019, Singletary2021, Origane2021, Ibuki2023}.
Then they were also applied to various problems, including maintaining connectivity \cite{Capelli2021, Fu2022, Luo2020}, coverage mission \cite{Ozkahraman2020}, energy autonomy \cite{Notomista2018, Fouad2022}, transporting payloads \cite{Hegde2021, Herguedas2022}, and herding animals that have strayed \cite{Sebastian2022}.
However, the effectiveness of CBFs in the control problem focused in this paper has not been fully studied.
Since this paper considers the connectivity aspects of a sensing network without relying on data transmission in the presence of obstacles—a departure from existing studies—we incorporate CBF constraints for preserving line-of-sight (LOS) while maintaining maximum and minimum distances between neighboring robots.
Another feature of our control problem is the difficulty of determining desired control inputs, given the dense and complex obstacle-laden environments under consideration, particularly because robots, except for the leader, do not know their destination.

In complicated control problems with multiple CBF constraints, optimization problems of control inputs are not necessarily feasible.
To deal with this problem, Breeden and Panagou \cite{Breeden2023} presented an iterative algorithm to obtain a set of CBFs which ensure that the set composed of those CBFs is a viability domain.
Molnar and Ames \cite{Molnar2023} proposed an algorithmic approach to formulate a single continuously differentiable CBF to capture complex safety specifications by multiple CBFs.
In this paper, we directly impose multiple CBF constraints in an optimization process, mirroring the approach taken in many other studies \cite{Rauscher2016, WangL2017, Luo2020, Chen2021, Fu2022, Fouad2022, Herguedas2022}.
The infeasibility of the optimization problems is addressed through a classical method using soft constraints \cite{Maciejowski2002}.
A possible future work is the incorporation of more sophisticated CBF construction algorithms, such as those outlined in \cite{Breeden2023, Molnar2023}, into our control method.

While these CBF-based methods resolve a numerical optimization problem at each time step, many control methods for multi-agent systems adopt simple control algorithms without solving numerical optimization \cite{Sakai2018, Mondal2018, Nomura2021, Qiao2022, Olfati2006, Su2009, Sakai2017}.
These methods determine control inputs through a simple combination of repulsive and attractive actions based on artificial potential fields (APFs) considering constraints among robots and obstacles.
However, as suggested in \cite{Singletary2021}, APF-based methods tend to suffer from more oscillations than CBF-based methods.
Thus, one important research direction is how to construct simple control algorithms without solving numerical optimizations while taking advantage of CBFs capable of generating smooth behaviors.
However, control algorithms that incorporate CBFs into these simple approaches for maneuvering through obstacle-laden environments have not been fully studied.

\subsection{Contribution}
In this paper, we propose a novel CBF-based control method for coordinating a group of robots in obstacle-laden environments while preserving the connectivity of the sensing network without data transmission between robots.
In this method, CBF constraints are constructed to preserve LOS integrity in addition to the maximum and minimum separation distances between neighboring robots.
This design is necessary to uphold the connectivity of a sensing network without data transmission in the presence of obstacles, unlike existing CBF control methods \cite{Ames2017, WangL2017, Chen2021, Ames2019, Singletary2021, Origane2021, Ibuki2023, Capelli2021, Fu2022, Luo2020, Ozkahraman2020, Notomista2018, Fouad2022, Hegde2021, Herguedas2022, Sebastian2022}.
Furthermore, unlike the existing method \cite{Nomura2021} based on APFs, the desired inputs are determined based on the CBF constraints to mitigate oscillatory movements and frequent constraint violations for UAVs.

In addition to the optimization-based method, we present an approximate method that does not rely on numerical optimization.
While CBFs are expected to improve oscillatory behaviors, they entail higher computational costs to solve numerical optimization problems.
Thus, it is practically important to investigate simple CBF-based algorithms without solving numerical optimization problems.

We perform extensive simulations to compare the performance of the proposed CBF-based methods against an APF-based method, considering aspects such as constraint violations, computation time, and oscillation of responses.
Simulation results show that both the optimization-based and approximate methods generate fewer oscillatory responses and constraint violations than an APF-based method.
Experimental results using real quadrotors are also shown.

A preliminary version of this work was presented in \cite{Thinesh2023}.
In this paper, we present a new approximate method that does not necessitate the iterative modification of input, as outlined in \cite{Thinesh2023}.
The optimization-based method is also introduced alongside intensive simulation results to compare the performance of the proposed CBF-based methods (the optimization-based and approximate methods) against an APF-based method.

\section{Problem formulation}\label{sec:problem}
We consider a group of \textit{N} homogeneous robots that operate in a 3D environment containing stationary obstacles.
Our control algorithm is constructed based on the assumption that the movement of the $i$th robot ($i=1,2,\ldots,N$) is described by the following second-order system:
\begin{align}
    \begin{aligned}
        \dot{x}_i & = v_i, \\
        \dot{v}_i & = u_i,\  \|u_i\| \leq \eta, \label{eq:sysdynamics}
    \end{aligned}
\end{align}
where $x_i \in \mathbb{R}^n, v_{i} \in \mathbb{R}^n$, and $u_i \in \mathbb{R}^n$ are the position, velocity, and control input of robot $i$, respectively.
The inclusion of a positive constant $\eta > 0$ in the input constraint accounts for hardware limitations.

We assume that a sufficient condition for collision avoidance between robots $i$ and $j$ is given as follows:
\begin{align}
    \| x_{i} - x_{j} \| \geq \underline{d}_c,~~~ \forall j \in \mathcal{V} \setminus \{i\}, \label{eq:coli1}
\end{align}
where $\mathcal{V} := \{1,2,\ldots, N \}$ is the set of indices of all robots.
Although a robot is modeled as a point in \eqref{eq:sysdynamics}, the parameter $\underline{d}_c$ in \eqref{eq:coli1} must be determined by taking into account the size of the actual robots.
We also assume that a sufficient condition for obstacle avoidance is given as
\begin{align}
    \| x_{i} - x_{o} \| \geq \underline{d}_o,~~~ \forall x_o \in \mathcal{O}, \label{eq:coli2}
\end{align}
where $\mathcal{O}$ is a set of all points on obstacles in the workspace.

To describe the sensing model, we first define the line segment joining $p$ and $q$ as
\begin{align}
    \mathcal{L}(p,q) := \{ (1-\lambda) p +  \lambda q , \forall \lambda \in [0,1] \}.
\end{align}
We then assume that robot $i$ can sense the position and velocity of robot $j \in \mathcal{V} \setminus \{i\}$ if the following conditions are satisfied
\begin{align}
     & \| x_j  - x_i  \| \leq d_s, \label{eq:range} \\
     & \| q - x_o \| \geq \underline{d}_{ls} ,\hspace{3mm} \forall q \in \mathcal{L}(x_i, x_j) ,\hspace{3mm} \forall x_o \in \mathcal{O}. \label{eq:los}
\end{align}
The condition in \eqref{eq:range} implies that the maximum sensing range is given by a positive number $d_s$.
The condition in \eqref{eq:los} implies that the distance from $\mathcal{L}(x_i, x_j)$ to each obstacle is at least the minimum clearance $\underline{d}_{ls}$, ensuring that obstacles do not interrupt the LOS between robots $i$ and $j$.
We also assume that robot $i$ can detect a point on an obstacle $x_o \in \mathcal{O}$, if
\begin{align}
     & \| x_o - x_i \| \leq d_s, \label{eq:range_o} \\
     & \| x_o - x_i \| \leq \| q - x_i \| ,\hspace{3mm} \forall q \in \bar{\mathcal{L}}(x_i, x_o) \cap \mathcal{O}, \label{eq:los_o}
\end{align}
where $\bar{\mathcal{L}}(x_i, x_o) := \{ (1-\lambda) x_i + \lambda x_o, \forall \lambda \geq 0 \}$.
While the set $\mathcal{L}(x_i, x_o)$ includes only points between $x_i$ and $x_o$, the set $\bar{\mathcal{L}}(x_i, x_o)$ includes points behind $x_o$ along the line from $x_i$ to $x_o$, in addition to the points in $\mathcal{L}(x_i, x_o)$.
Thus, the condition in \eqref{eq:los_o} implies that no other point on an obstacle is closer to $x_i$ than $x_o$ on $\mathcal{L}(x_i, x_o)$.
We denote $\mathcal{O}_i (t)$ as the set of points on obstacles detected by robot $i$ at time $t$.

In terms of the sensing described above, we represent the network topology of the multi-robot system using a graph $\mathcal{G}_s(x(t)) = (\mathcal{V}, \mathcal{E}_s(x(t)))$ where $x = [x_1, x_2, \ldots, x_N]$.
We denote $\mathcal{V} $ and $\mathcal{E}_s (x(t))$ as the node set and edge set, respectively.
The elements of $\mathcal{E}_s (x(t))$ are pairs of robot indices that can sense each other's positions at time $t$.
A connected graph $\mathcal{G}_s$ implies that, for every pair of nodes, there exists a path from one node to the other.

We assume that a target path is given only to the leader of the $N$ robots, whose index is set as $N$ without loss of generality.
The remaining robots $i$ ($=1, 2, \ldots, N-1$) are called followers.
Each follower is presumed to be unaware of whether another robot serves as the leader.
If $\mathcal{G}_s$ is connected, a pathway exists between the leader and each follower.
Thus, since the maximum length of each link of $\mathcal{G}_s$ is kept to no more than a given finite value $d_s$, each follower is forced to follow the leader at a certain distance (at most $(N-1) d_s$).
Furthermore, reducing the length of the link to below $d_s$ allows for a decrease in the distance between the leader and a follower.
Thus, in this paper we aim to preserve the connectivity of the following subgraph of $\mathcal{G}_s(x(t))$,
\begin{align}
     & \mathcal{G}_m(x(t)) := (\mathcal{V}, \mathcal{E}_m(x(t))), \label{eq:calGn}                                 \\
     & \mathcal{E}_m(x(t)) := \{(i,j) \in \mathcal{E}_s(x(t) ) ~|~ \| x_i(t) - x_j(t) \| \leq \bar{d}_m \}, \notag
\end{align}
where $\underline{d}_c < \bar{d}_m < d_s$.
Then, the neighbors of robot $i$ are defined as
\begin{align}
    \mathcal{N}_i(x(t)) := \{j ~ |~ (i,j) \in \mathcal{E}_m(x(t))\}.
\end{align}
From the definition of $\mathcal{G}_m(x(t))$, the connectivity of $\mathcal{G}_s$ is preserved if that of $\mathcal{G}_m(x(t))$ is preserved.

We assume that $\mathcal{G}_m $ is connected at the initial time $t=0$.
Thus, the simplest way to preserve connectivity is to control the robots such that the edges of $\mathcal{G}_m$ at $t=0$ are not lost.
However, in obstacle-laden environments, it is often necessary to change the network topology to facilitate navigation through a narrow space.
Thus, it is necessary to select edges to be maintained such that the connectivity of $\mathcal{G}_m $ is preserved.
To describe the edges to be preserved, we define the symmetric indicator function $\sigma_{ij}(t) = \sigma_{ji}(t) \in \{0,1\}$.
If $\sigma_{ij} = 1$, there will be an effort to preserve the edge $(i,j)$.
Essentially, robot $i$ aims to preserve the link to robot $j$ in the following set:
\begin{align}
    \mathcal{N}_i^{\sigma}(x(t)) := \{j \in \mathcal{N}_i(t)  ~|~  \sigma_{ij}(t)=1 \}.
\end{align}
We also define the following subgraph of $\mathcal{G}_m(x(t))$ as
\begin{align}
     & \mathcal{G}_{\sigma}(x(t)) := (\mathcal{V}, \mathcal{E}_{\sigma}(x(t))), \\
     & \mathcal{E}_{\sigma}(x(t)) := \{(i,j) \in \mathcal{E}_m(x(t) )  ~|~ \sigma_{ij}(t)=1 \}. \label{eq:Esigma}
\end{align}
By ensuring that $\sigma_{ij}$ is set in a manner that renders $\mathcal{G}_{\sigma}$ connected, the connectivity of $\mathcal{G}_m$ is maintained by orchestrating the movement of the robots to avoid any loss of edges within $\mathcal{G}_{\sigma}$.

We propose an algorithm to compute $u_i$ in a manner that upholds the connectivity of $\mathcal{G}_m$, while adhering to the collision avoidance conditions in \eqref{eq:coli1} and \eqref{eq:coli2}.
However, when applied to UAVs, it is difficult to avoid constraint violations completely due to modeling errors and disturbances, especially when multiple constraints are imposed severely.
Thus, for the constants satisfying $d_m < \bar{d}_m$, $d_c > \underline{d}_c$, $d_o > \underline{d}_o$, and $d_{ls} > \underline{d}_{ls}$, we control the robots so as to satisfy the following constraints:
\begin{enumerate}[label=\roman*)]
    \item The maximum distance constraint
          \begin{align}
              \| x_i  - x_j  \| \leq d_m, ~~~ \forall j \in \mathcal{N}_i^{\sigma}(t). \label{eq:c1}
          \end{align}

    \item The inter-robot collision avoidance constraint
          \begin{align}
              \| x_i  - x_j  \| \geq d_c, ~~~ \forall j \in \mathcal{N}_i(t). \label{eq:c2}
          \end{align}

    \item The obstacle avoidance constraint
          \begin{align}
              \| x_i  - x_o  \| \geq d_o, ~~~ \forall  x_o \in \mathcal{O}_i(t). \label{eq:c3}
          \end{align}

    \item The LOS preservation constraint for each $j \in \mathcal{N}_i^{\sigma}(t)$
          \begin{align}
              \| q - x_o \| \geq d_{ls} ,\hspace{3mm} \forall x_o \in \mathcal{O}, \hspace{3mm} \forall q \in \mathcal{L}(x_i, x_j). \label{eq:c4}
          \end{align}
\end{enumerate}
If all of these constraints are satisfied at time $t$, the state is referred to as Normal mode.
On the other hand, if one or more constraints are violated at $t$, we use a different control law to recover from the constraint violation, referred to as Recovery mode.
In this paper, we focus mainly on Normal mode and propose new control methods based on CBFs.

\section{Constraints based on CBFs} \label{sec:apf_cbf}

\subsection{Motivation} \label{sec:cbf_moti}
To explain the motivation to use CBF-based constraints, we describe a limitation of the APF-based approach used in \cite{Sakai2018, Nomura2021}.

\begin{figure}
    \centering
    \includegraphics[width=0.65\linewidth]{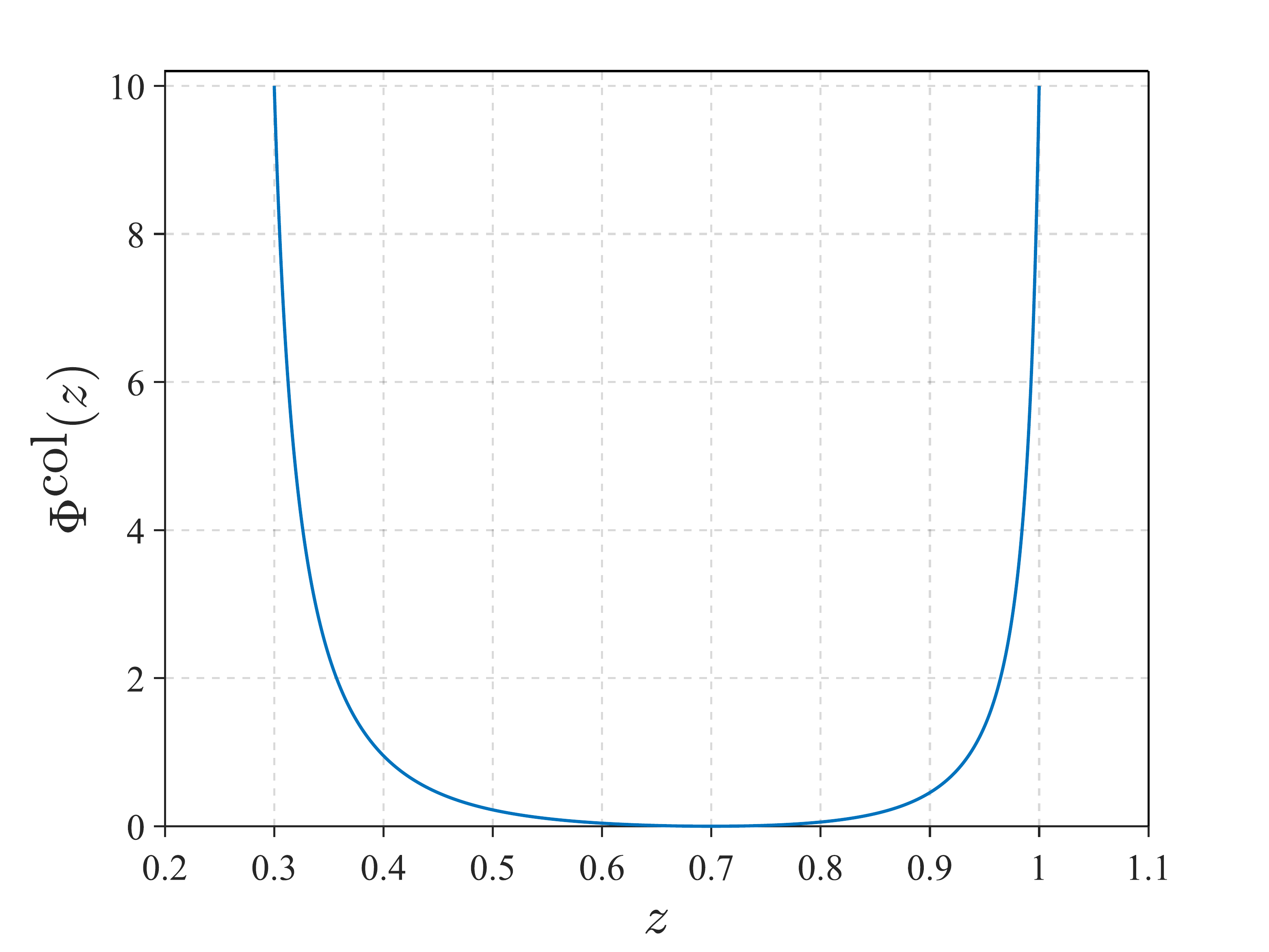}
    \caption{Example of $\Phi^{\rm col}(z) $ ($d_c = 0.3, d_r = 0.7, d_m = 1.0$).}
    \label{fig:J1}
\end{figure}

For the constraints in \eqref{eq:c1} and \eqref{eq:c2}, the following repulsive APF is used in \cite{Sakai2018, Nomura2021},
\begin{align}
     \Psi_{i}^{\rm col}(x)  &= \sum_{j  \in \mathcal{N}_i^{\sigma}} \Phi^{\rm col} (\| x_i  - x_j \|) \\
     \Phi^{\rm col}(z)      &:= \frac{(z - d_r) ^2 (d_m - z) }{ (d_m - d_c)^2 (z - d_c) + (d_r - d_c) ^2 (d_m - z) / \kappa_1 } \nonumber \\
                            &+ \frac{ (z - d_c) (z - d_r)^2 }{ (d_m - d_c)^2 (d_m - z) + (z - d_c) (d_m - d_r) ^2 / \kappa_2} \notag
\end{align}
where $\kappa_1$ and $\kappa_2$ are design parameters whose values are equivalent to $\Phi^{\rm col}(z)$ at $z=d_c$ and $z=d_m$, respectively.
Figure \ref{fig:J1} illustrates an example of $\Phi^{\rm col}(z)$ for $d_c = 0.3, d_r = 0.7, d_m = 1.0, \kappa_1 = \kappa_2 = 10$.
As the example shows, $\Phi^{\rm col}(z)$ at the minimum value at the desired relative distance $d_r$, and monotonically increases as $z$ approaches the maximum allowable distance $d_m$ or the minimum allowable distance $d_c$.
Similarly, for the constraints in \eqref{eq:c3} and \eqref{eq:c4}, repulsive APFs increase as the distance decreases towards the minimum limits $d_o$ and $d_{ls}$, respectively.

A limitation of the APF-based approach is that almost no action is taken when the distance is far from the maximum and minimum limits, irrespective of how high the robot's speed in advancing towards a limit.
As a result, if the velocity aimed at reaching a constraint boundary becomes too high, the constraint is violated even if the maximum allowable repulsive acceleration is applied near the constraint boundary to avoid violation.
Furthermore, the repulsive acceleration is large when the distance is close to the maximum and minimum limits, irrespective of how slow the robot is moving towards the constraint boundary.
Thus, when the robot approaches a constraint boundary slowly, an excessively strong repulsive force is applied, which causes oscillatory behaviors.
Although a damping filter was used to suppress oscillations in \cite{Nomura2021}, this approach does not address the root cause of the problem.
In other words, such a damping filter increases the risk of constraint violations by potentially eliminating essential actions required to avoid constraint violations.
On the other hand, CBF-based methods consider velocity constraints depending on the proximity to their limits, as described in Section \ref{sec:cbf_const}.

\subsection{Formulation of CBF constraints} \label{sec:cbf_const}
For a constraint $h(x) \geq 0$ to the position vector of robots, $x:=[x_1,x_2,\ldots,x_N]^T$, CBF-based methods consider the following derivative condition,
\begin{align}
    \dot{h}(x,v) = \nabla h(x)^T \dot{x} = \nabla h(x)^T v \geq -\kappa(h(x)) \label{eq:h_dot}
\end{align}
where $\kappa(\cdot)$ is an extended class $\mathcal{K}$ function.
In other words, the lower limit of $\dot{h}(x,v)$ is 0 when the constraint reaches its minimum limit, $h(x)=0$, and diminishes further as the distance from the limit increases.
Thus, this constraint aims to prevent high speeds of $h(x)$ towards the limit, $h(x)=0$.

The constraint in \eqref{eq:c2} for collision avoidance between robots can be described as
\begin{align}
    h_{ij}^c(x) := \|x_i - x_j\| - d_c \geq 0,\ \quad \forall j \in \mathcal{N}_i. \label{eq:hc}
\end{align}
To keep this constraint, we can obtain the following derivative condition in a manner similar to \eqref{eq:h_dot} for $\kappa (h_{ij}^c) = \alpha_c h_{ij}^c$. 
\begin{align}
    G_{ij}^c(x) := {\dot{h}}_{ij}^c(x) + \alpha_c h_{ij}^c(x) \geq0, \  \forall j \in \mathcal{N}_i, \label{eq:Gc}
\end{align}
where
\begin{align}
    {\dot{h}}_{ij}^c(x)  = \left(\mathrm{\nabla}_{x_i}h_{ij}^c(x)\right)^T v_i + \left(\mathrm{\nabla}_{x_j}h_{ij}^c(x)\right)^T v_j.  \label{eq:hc_dot}
\end{align}

Several approaches \cite{Nguyen2016, WangL2017, Xiao2019, Ibuki2023} are available to deal with derivative conditions which do not include the control input as in \eqref{eq:Gc}.
In this paper, we adopt the approach in \cite{WangL2017}, considering a safety margin determined by the maximum braking acceleration, as described below.
Comparing the control performance with other approaches is a potential area for future exploration.

Considering a safety margin based on the maximum relative deceleration $2\eta$, the constraint in \eqref{eq:hc} is modified as
\begin{align}
    h_{ij}^c(x) + \bar{v}_{ij}(t) T_{ij} + 2 \int_{0}^{T_{ij}} \eta \tau d\tau \geq 0 \label{eq:max_brake}
\end{align}
where $\bar{v}_{ij}(t):=\frac{x_{ij}^T(t)}{\| x_{ij}(t) \|} v_{ij}(t)$, $x_{ij} := x_i - x_j$, $v_{ij} := v_i - v_j$, and $T_{ij} := \frac{0 - \bar{v}_{ij}(t)}{2\eta}$.

When both robots $i$ and $j$ apply the maximum deceleration, the relative velocity $\bar{v}_{ij}(t+T_{ij}) = \bar{v}_{ij}(t)-2\eta T_{ij}$ diminishes to 0.
From \eqref{eq:max_brake}, a constraint $\bar{h}_{ij}^c(x,v) \geq 0$ is obtained, where
\begin{align}
    \bar{h}_{ij}^c(x,v) := \sqrt{4\eta h_{ij}^c(x)} + \frac{x_{ij}^T(t)}{\| x_{ij}(t) \|} v_{ij}(t). \label{eq:hc_bar}
\end{align}
Then, a derivative condition can be derived for \eqref{eq:hc_bar} as follows,
\begin{align}
    x_{ij}^T(t) u_{ij}(t) + b_{ij}^c(t) \geq 0, \label{eq:hc_bar_dot}
\end{align}
where
\begin{align}
    b_{ij}^c := \alpha_c (\bar{h}_{ij}^c)^3 \|x_{ij}\| - (\bar{v}_{ij})^2 + \|v_{ij}\|^2 + \sqrt{\frac{\eta}{h_{ij}^c}}  v_{ij}^T x_{ij}. \label{eq:b_ij_c}
\end{align}
Based on \eqref{eq:hc_bar_dot}, the following distributed constraint is proposed in \cite{WangL2017}.
\begin{align}
    x_{ij}^T(t) u_{i}(t) + b_{ij}^c(t)/2  & \geq 0, \\
    -x_{ij}^T(t) u_{j}(t) + b_{ij}^c(t)/2 & \geq 0. \label{eq:Gc_bar}
\end{align}
Thus, the constraint for $u_i$ to adhere to the constraint in \eqref{eq:c2} is described as
\begin{align}
    A_{ij}^c u_{i}(t) + b_{ij}^c(t)/2 \geq 0, \quad \forall j \in \mathcal{N}_i(t) \label{eq:Gc_bar_distributed}
\end{align}
where $A_{ij}^c := x_{ij}^T(t)$.
Similarly, a CBF-based constraint to maintain the constraint in \eqref{eq:c1} is described as
\begin{align}
    G_{ij}^m(x) := {\dot{h}}_{ij}^m(x) + \alpha_m h_{ij}^m(x) \geq0,\  \forall j \in \mathcal{N}_i^\sigma, \label{eq:Gm}
\end{align}
for $ h_{ij}^m(x) := d_m - \|x_i - x_j\|$ and a positive constant $\alpha_m$.
A similar constraint to \eqref{eq:Gc_bar_distributed} is derived as
\begin{align}
    A_{ij}^m u_{i}(t) + b_{ij}^m(t)/2 \geq 0, \quad \forall j \in \mathcal{N}_i^\sigma(t), \label{eq:Gm_bar_distributed}
\end{align}
where $A_{ij}^m := -x_{ij}^T(t)$ and
\begin{align}
    b_{ij}^m         & := \alpha_m (\bar{h}_{ij}^m(x))^3 \|x_{ij}\| + (\bar{v}_{ij})^2 - \|v_{ij}\|^2 -\sqrt{\frac{\eta}{h_{ij}^m }}  v_{ij}^T x_{ij}, \notag \\
    \bar{h}_{ij}^{m} & := \sqrt{4\eta h_{ij}^m(x)} - \frac{x_{ij}^T(t)}{\| x_{ij}(t) \|} v_{ij}(t). \label{eq:hm_bar}
\end{align}
For the obstacle avoidance constraint in \eqref{eq:c3}, a CBF-based constraint is described as
\begin{align}
    G_{io}^{ob}(x_i,x_o) := {\dot{h}}_{io}^{ob}(x_i,x_o) + \alpha_{ob} h_{io}^{ob}(x_i,x_o) \geq0, \label{eq:Go} 
\end{align}
for $ h_{io}^{ob}(x) := \|x_i - x_o\| - d_o$ and a positive constant $\alpha_{ob}$.
A modified constraint including $u_i$ is obtained as
\begin{align}
    A_{io}^{ob} u_{i}(t) + b_{io}^{ob}(t) \geq 0, \label{eq:Go_bar}
\end{align}
where $A_{io}^{ob} := x_{io}^T(t) $, $x_{io}:= x_i - x_o$, and
\begin{align}
    b_{io}^{ob}       & := \alpha_{ob} (\bar{h}_{io}^{ob})^3 \|x_{io}\| - \frac{(v_{i}^T x_{io})^2}{\|x_{io}\|^2} + \|v_{i}\|^2 + \sqrt{\frac{2\eta}{h_{io}^{ob}}}  v_{i}^T x_{io}, \notag \\
    \bar{h}_{io}^{ob} & := \sqrt{2\eta h_{io}^{ob}(x)} + \frac{x_{io}^T(t)}{\|x_{io}(t)\|} v_{i}(t). \label{eq:ho_bar}
\end{align}

For implementation, it is not realistic to consider the constraint in \eqref{eq:Go_bar} for each $x_o \in \mathcal{O}_i(t)$.
Therefore, we approximate $\mathcal{O}_i(t)$ by a union of convex sets such as lines and ellipsoids, and consider the constraints in \eqref{eq:Go_bar} for the nearest point $x_o^*$ in each convex set.

In Section \ref{sec:cbf_los}, we derive a CBF-based constraint to keep the constraint in \eqref{eq:c4} for LOS preservation.

\subsection{Constraints for LOS preservation} \label{sec:cbf_los}
In the same way as in the case of \eqref{eq:Go_bar}, we approximate $\mathcal{O}_i(t)$ through a union of convex sets such as lines and ellipsoids, deriving a CBF-based constraint for the nearest point from the LOS line, and $x_o^*$, in each convex set.

\begin{figure}
    \centering
    \includegraphics[width=0.7\linewidth]{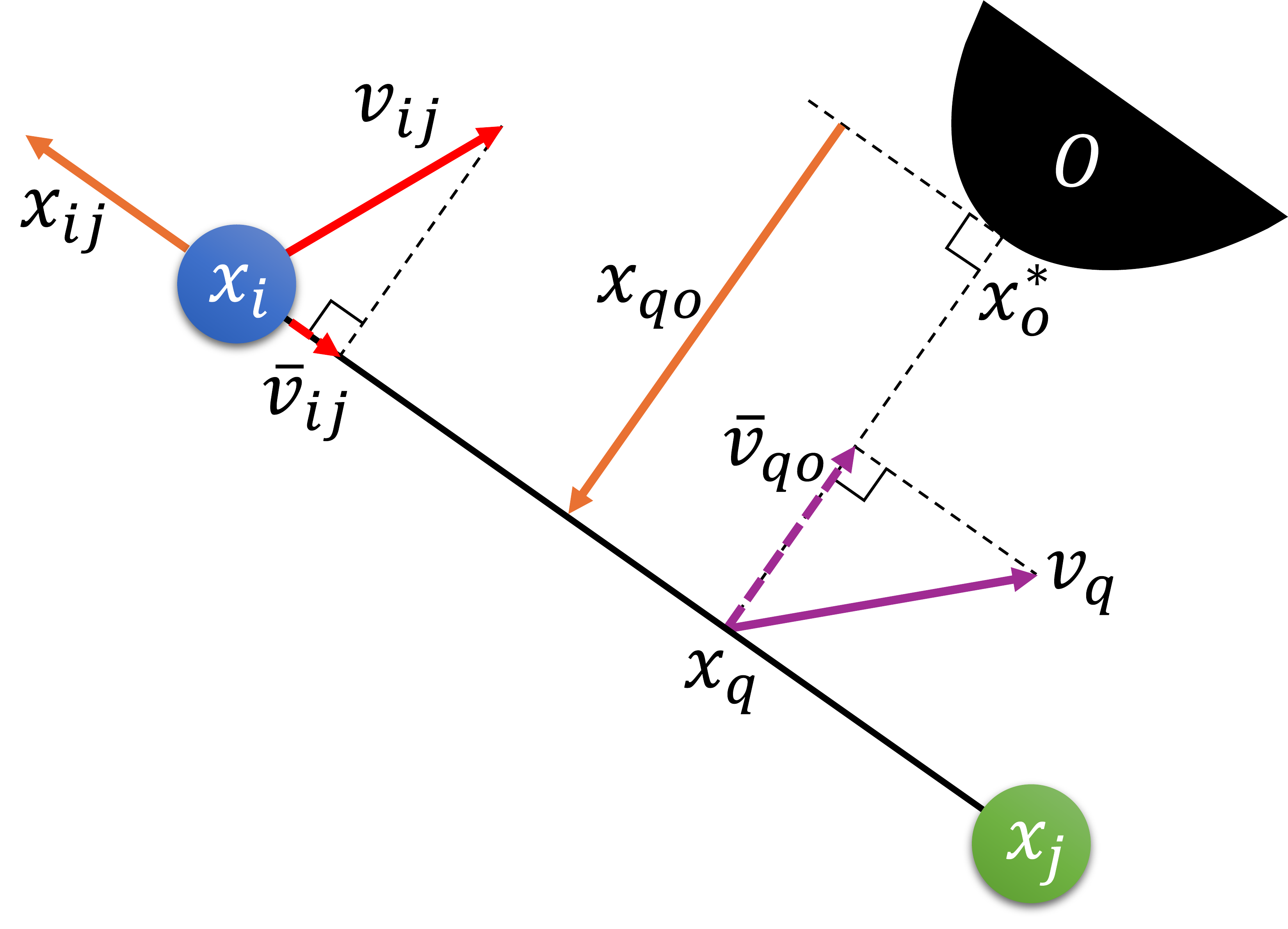}
    \caption{Relative position and velocity between robots $i$ and $j$, and illustration of point $q$ on LOS.}
    \label{fig:Relative_pos_vel}
\end{figure}

The position $x_q$ of the point $q$ projected from $x_o^*$ onto the LOS between $i$ and $j$, as shown in Figure \ref{fig:Relative_pos_vel}, is described as
\begin{align}
    x_{q} := (1-\lambda^*) x_i + \lambda^* x_j,~~~ \lambda^* := \frac{(x_{i} - x_o^*)^T x_{ij}}{\|x_{ij}\|^2}.
\end{align}
For $x_{qo} := x_q - x_o^*$, $h_{ijo}^{ls}(x,x_o):= \|x_{qo}\| - d_{ls}$, and a positive number $\alpha_{ls}$,
we consider the following CBF-based constraint,
\begin{align}
    G_{ijo}^{ls}(x,x_o) & := {\dot{h}}_{ijo}^{ls} + \alpha_{ls} h_{ijo}^{ls} \geq 0, \quad \forall j \in \mathcal{N}_i^\sigma. \label{eq:Gls}
\end{align}
We define the velocity of point $q$ as
\begin{align}
    v_{q} := (1-\lambda^*) v_i + \lambda^* v_j,
\end{align}
and its projection onto $x_{qo}$ as
\begin{align}
    \bar{v}_{qo}(t) := \frac{x_{qo}^T(t)}{\|x_{qo}(t)\|} v_{q}(t).
\end{align}
In the same way as in \eqref{eq:max_brake}, we consider the following constraint accounting for a safety margin in a scenario where both robots $i$ and $j$ change the velocity of the LOS towards $x_o^*$ from $\bar{v}_{qo}$ to $0$ using the maximum deceleration $\eta$.
\begin{align}
    h_{ijo}^{ls} + \bar{v}_{qo}(t) T_{qo} + \int_{0}^{T_{qo}} \eta \tau d\tau \geq 0, \label{eq:max_brake_los}
\end{align}
where $T_{qo}:=\frac{0-\bar{v}_{qo}(t)}{\eta}$.
From \eqref{eq:max_brake_los}, a constraint $\bar{h}_{ijo}^{ls}(x,v) \geq 0$ is obtained, where
\begin{align}
    \bar{h}_{ijo}^{ls} := \sqrt{2\eta h_{ijo}^{ls}} + \frac{x_{qo}^T(t)}{\| x_{qo}(t) \|} v_{q}(t). \label{eq:hls_bar}
\end{align}
A derivative condition can be derived for \eqref{eq:hls_bar} as follows,
\begin{align}
    \left(\nabla_{v_i} \bar{h}_{ijo}^{ls} \right)^T u_{i}(t) + \left(\nabla_{v_j} \bar{h}_{ijo}^{ls} \right)^T u_{j}(t) + b_{ijo}^{ls}(t) \geq 0, \label{eq:Gls_bar}
\end{align}
where $b_{ij}^c := \left(\nabla_{x_i} \bar{h}_{ijo}^{ls}\right)^T v_{i}(t) + \left(\nabla_{x_j} \bar{h}_{ijo}^{ls}\right)^T v_{j}(t) + \alpha_{ls} (\bar{h}_{ij}^{ls})^3$.
%
The constraint in \eqref{eq:Gls_bar} is distributed as follows.
\begin{align}
    \left(\nabla_{v_i} \bar{h}_{ijo}^{ls} \right)^T u_{i}(t) + \lambda^* b_{ij}^c(t)     &\geq 0, \\
    \left(\nabla_{v_j} \bar{h}_{ijo}^{ls} \right)^T u_{j}(t) + (1-\lambda^*) b_{ij}^c(t) &\geq 0. \label{eq:hdot_ij_c_dist}
\end{align}
Thus, the constraint for robot $i$ to maintain the LOS preservation constraint in \eqref{eq:c4} is described as
\begin{align}
    G_{ij}^{ls}(x) := A_{ij}^{ls} u_{i}(t) + \lambda^* b_{ij}^{ls}(t) \geq 0 \label{eq:Gls_bar_distributed}
\end{align}
where $A_{ij}^{ls} := \left( \nabla_{v_i} \bar{h}_{ijo}^{ls} \right)^T$.

\section{Control algorithm} \label{sec:ctrl_alg}
The outline of the control algorithm for robot $i$ is described as follows.\\
Step 1: The indicator function $\sigma_{ij}(t)$ for all $j \in \mathcal{N}_i(t)$ is determined to select links in $\mathcal{{G}}_m$ to be preserved.\\
Step 2: The desired control input is determined by considering the CBF-based constraints and APFs.\\
Step 3: If the desired control input violates CBF-based constraints, it is minimally modified to obtain the final control input $u_i$.\\
Each step of the algorithm is described in detail in the following subsections.

\subsection{Link management (Step 1)} \label{sec:ctrl_alg:s1}
The purpose of link management is to achieve a slender swarm shape with fewer links between robots when navigating narrow passages and to enable the swarm to navigate through dense obstacle-laden environments.
To achieve a slender shape with fewer links between robots, it is necessary to deactivate some of the redundant edges in the formation graph $\mathcal{G}_m$.
To avoid leaving some robots behind, this deactivation process needs to be performed while preserving connectivity.
In this paper, we adopt simple rules proposed in \cite{Sakai2018, Nomura2021} to deactivate edges in a decentralized manner without data transmission, as described below.
See \cite{Sakai2018} for theoretical analysis on connectivity preservation of $\mathcal{G}_\sigma$ after these decentralized deactivation rules are applied.

\begin{figure}
    \centering
    \includegraphics[width=0.5\linewidth]{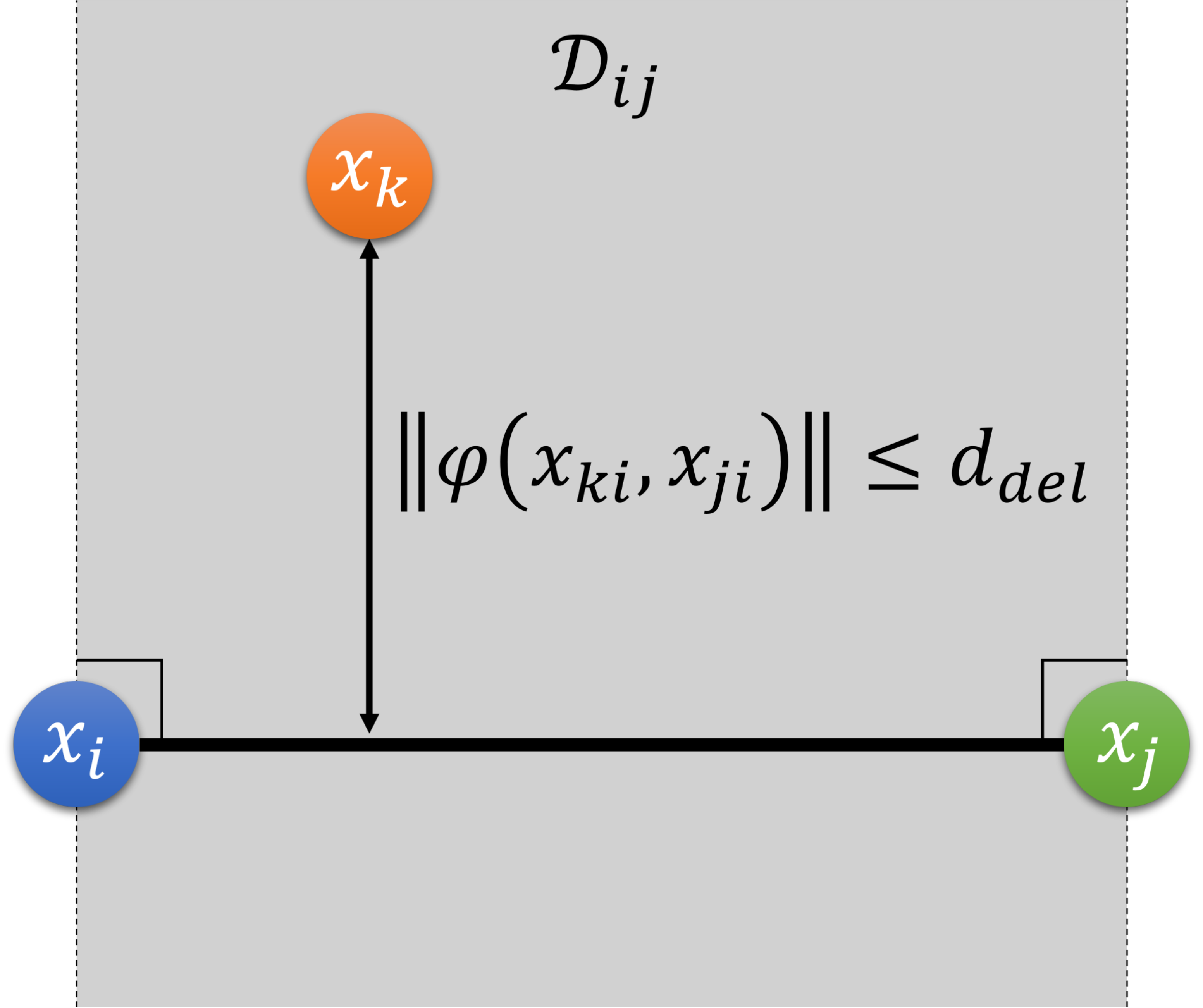}
    \caption{2D illustration of $D_{ij}$ and $\|\varphi(x_{ki}, x_{ji})\|$.}
    \label{fig:D_ij}
\end{figure}

The region $\mathcal{D}_{ij}$ is defined for $x_i$ and $x_j$, as shown in Figure \ref{fig:D_ij}, to describe the link deactivation rule,
\begin{align}
    \mathcal{D}_{ij} := \{q ~|~ (q - x_i)^T x_{ji} > 0,\ (q - x_j)^T x_{ji} < 0\}. \label{eq:Dij}
\end{align}
Furthermore, a projection function $\varphi(p,q)$ is employed to project the vector $p$ onto the plane perpendicular to vector $q$.
The projection function can be defined as follows,
\begin{align}
    \varphi(p,q) := p - \frac{p^Tq}{\|q\|^2}q.
\end{align}
As shown in Figure \ref{fig:D_ij}, the distance between $x_k$ and the line connecting $x_i$ and $x_j$ can be computed as $\|\varphi(x_{ki}, x_{ji})\|$.
If robots $i, j$, and $k$ satisfy the following condition, then the link $(i,j) \in \mathcal{E}_m$ is deactivated:
\begin{align}
    \begin{aligned}
         & \|\varphi(x_{ki},x_{ji})\| \leq d_{del},\  x_k \in \mathcal{D}_{ij}, \\
         & (i,j) \in \mathcal{E}_m,\  (j,k) \in \mathcal{E}_m,\  (k,i) \in \mathcal{E}_m, \label{eq:sigma_c1}
    \end{aligned}
\end{align}
where $d_{del}$ is a positive constant satisfying $d_{del} < d_c \sin{\frac{\pi}{3}}$.

Another rule is that if only one edge of a connected robot triangle has a length of $d_m$, then that edge will be deactivated.
Specifically, the link $(i,j)$ will be deactivated if robots $i, j$, and $k$ satisfy the following condition,
\begin{align}
    \begin{aligned}
         & d_m - \delta_m < \|x_{ji}\| \leq d_m,\ \|x_{jk}\| < d_m - \delta_m, \\
         & \|x_{ki}\| < d_m - \delta_m,\ (i,j) \in \mathcal{E}_m,\ (j,k) \in \mathcal{E}_m,\ (k,i) \in \mathcal{E}_m, \label{eq:sigma_c2}
    \end{aligned}
\end{align}
where $\delta_m$ is a small positive number relative to $d_m$.
In summary, $\sigma_{ij}$ is defined as follows:
\begin{align}
    \sigma_{ij} = \begin{cases}
                      0, & \text{if } \eqref{eq:sigma_c1} \text{ or } \eqref{eq:sigma_c2} \\
                      1, & \text{otherwise}
                  \end{cases}.
\end{align}

\subsection{Desired control input calculation (Step 2)} \label{sec:ctrl_alg:s2}
Since the leader is given a target path, the desired control input for the leader is determined to follow that path.
In simulations and experiments, the following simple controller is used for a sequence of waypoints on the target path:
\begin{align}
    a_N = k_p(x_{wp} - x_N), \ k_p > 0,
\end{align}
where $x_{wp}$ is the next waypoint on the target path.

Determining desired inputs for the follower robots is more challenging because they do not have knowledge of the target path.
The desired control input for follower robots, $a_i$, is expressed by the following equation,
\begin{align}
    a_{i} = a_{i}^{con} + a_{i}^{da} + a_{i}^{ag}, \label{eq:a_des}
\end{align}
where $a_i^{con}$ has four components as follows,
\begin{align}
    a_{i}^{con} = a_{i}^{m} + a_{i}^{c} + a_{i}^{ob} + a_{i}^{ls}. \label{eq:a_con}
\end{align}
The significance of each component in \eqref{eq:a_des} and \eqref{eq:a_con} is elucidated in Table \ref{table:acc_description}.

\begin{table}[htbp]
    \renewcommand{\arraystretch}{1.5}
    \tbl{Description of Acceleration Components.}
    {\begin{tabular}{|c|l|}
            \hline
            $a_{i}^{m}$  & Acceleration to enforce maximum link length        \\ \hline
            $a_{i}^{c}$  & Acceleration to avoid collision with another robot \\ \hline
            $a_{i}^{ob}$ & Acceleration to avoid collision with obstacle      \\ \hline
            $a_{i}^{ls}$ & Acceleration to preserve LOS between robots        \\ \hline
            $a_{i}^{da}$ & Acceleration to avoid deadlock                     \\ \hline
            $a_{i}^{ag}$ & Acceleration to make the group more cohesive       \\ \hline
        \end{tabular}}
    \label{table:acc_description}
    \renewcommand{\arraystretch}{1}
\end{table}

In the previous work \cite{Nomura2021}, desired inputs were determined by employing repulsive APFs based on specified constraints and an attractive APF to foster group cohesion.
However, the repulsive APFs caused oscillatory behaviors in the desired inputs, as mentioned in Section \ref{sec:cbf_moti}.
In this paper, we use repulsive functions based on CBFs to compute $a_{i}^{con}$ instead of APFs.

\begin{figure}
    \centering
    \includegraphics[width=0.8\linewidth]{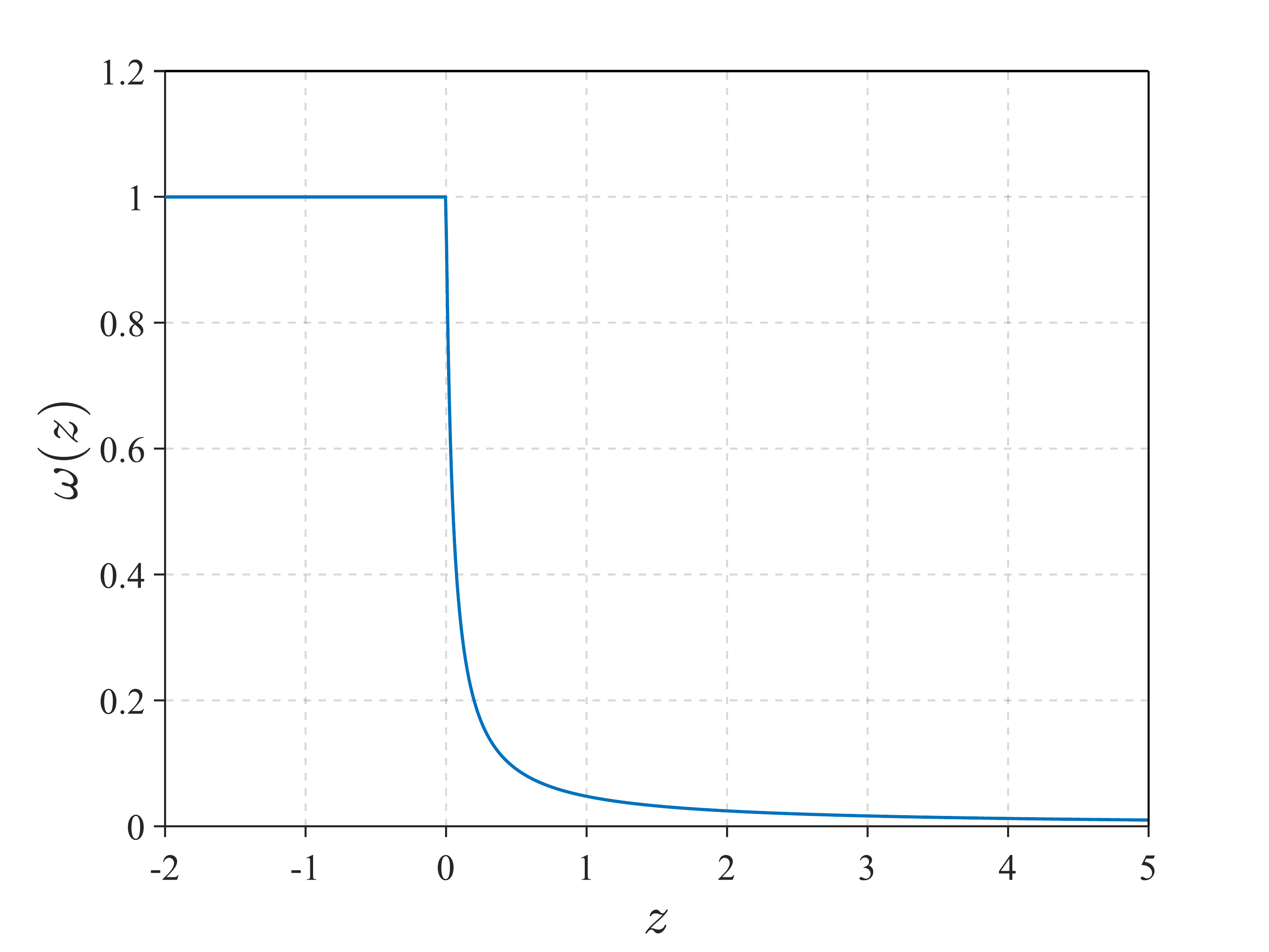}
    \caption{Example of $\omega_{m}(z)$ for $\mu_m = 1, \beta_m = 0.1$.}
    \label{fig:omega}
\end{figure}

The acceleration component $a_{i}^{m}$ is determined using the following repulsive force to satisfy the constraint in \eqref{eq:Gm}
\begin{align}
    a_{i}^{m} = \frac{1}{n_i^\sigma} \sum_{j\in\mathcal{N}_i^\sigma}{\omega_m(G_{ij}^m(x))\ \nabla_{v_i} G_{ij}^m(x)},
\end{align}
where $n_i^\sigma$ denotes the cardinality of the set $\mathcal{N}_i^\sigma$, and ${\omega}_m$ is defined as
\begin{align}
    \omega_{m}(z) := \frac{\mu_m\ \beta_m}{|z| + z + \beta_m},\ \mu_m > 0, \ \beta_{m} > 0. \label{eq:acc_weight}
\end{align}
Figure \ref{fig:omega} illustrates an example of the function $\omega_{m}(z)$ for $\mu_m = 1, \beta_m = 0.1$.
Here, the variable $\mu_m$ establishes the maximum value of the function, while $\beta_m$ determines the rate at which $\omega_{m}(z)$ decreases for $z \geq 0$.
Similarly, $a_{i}^{c}$ is determined through the following repulsive force to satisfy \eqref{eq:Gc}
\begin{align}
    a_{i}^{c} = \frac{1}{n_i} \sum_{j\in\mathcal{N}_i}{\omega_c(G_{ij}^c(x))\ \nabla_{v_i} G_{ij}^c(x)}, 
\end{align}
using $\omega_c(z)$ defined in the same way as \eqref{eq:acc_weight} for constants $\mu_c$ and $\beta_c$. $n_i$ denotes the cardinality of set $\mathcal{N}_i$.

To define $a_{i}^{ob}$, we specify the set $\mathcal{O}_i^{ob}$ comprising the nearest obstacle points detected from robot $i$, 
\begin{align}
    \mathcal{O}_i^{ob} = \operatorname*{argmin}_{x_o\in\mathcal{O}_i}{ \|x_i - x_o\| }.
\end{align}
The accelerations to avoid collision with obstacles is determined by:
\begin{align}
    a_{i}^{ob} = \frac{1}{n_i^{ob}} \sum_{x_o\in \mathcal{O}_i^{ob}}{\omega_{ob}(G_{io}^{ob}(x_i,x_o))\ \nabla_{v_i} G_{io}^{ob}(x_i,x_o)}.
\end{align}
where $\omega_{ob}(z)$ is defined in the same way as in \eqref{eq:acc_weight} for constants $\mu_o$ and $\beta_o$.

To elaborate on the term $a_{i}^{ls}$, we define the set $\mathcal{LS}_i$, which represents the neighboring robot and the corresponding obstacle that are nearest to the boundary of the LOS preservation condition in \eqref{eq:c4}, defined as follows:
\begin{align}
    \mathcal{LS}_i = \operatorname*{argmin}_{j\in\mathcal{N}_i^\sigma, \ x_o\in\mathcal{O}_i\cap\mathcal{D}_{ij}}{\|\varphi(x_{oi},x_{ji})\| }
\end{align}
The acceleration to preserve LOS is determined by the following equation,
\begin{align}
    a_{i}^{ls} & = \frac{1}{n_i^{ls}} \sum_{(j,x_o)\in{\mathcal{LS}}_i}{\omega_{ls}(G_{ijo}^{ls}(x,x_o))\ \nabla_{v_i} G_{ijo}^{ls}(x,x_o)},
\end{align}

\begin{figure}
    \centering
    \includegraphics[width=0.5\linewidth]{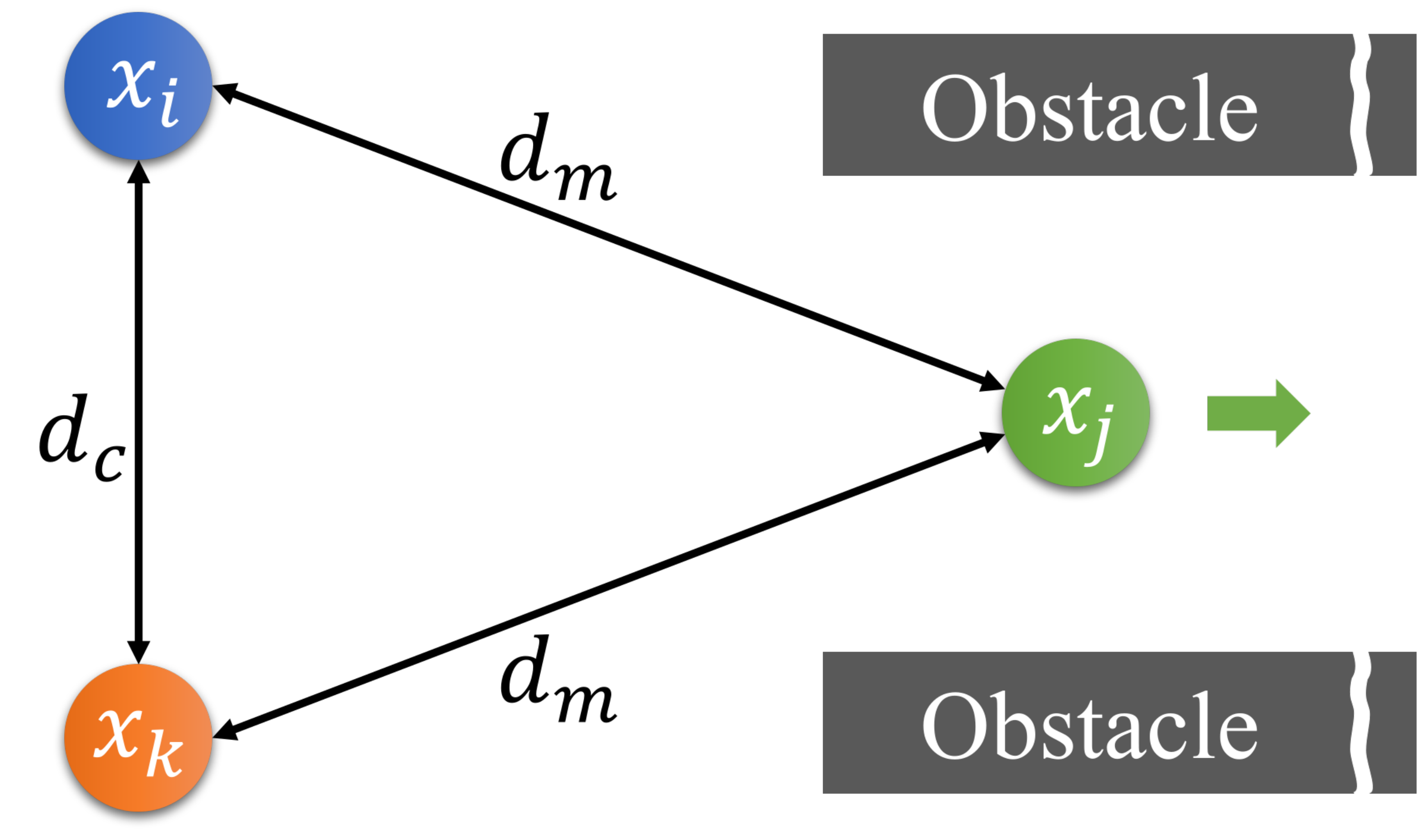}
    \caption{2D illustration of deadlock isosceles triangle.}
    \label{fig:Deadlock}
\end{figure}

A deadlock situation may occur when three robots forming an isosceles triangle move into a narrow space, as illustrated in Figure \ref{fig:Deadlock}.
In this scenario, the robots cannot find the correct link to deactivate without losing connectivity; thus the group will get stuck without being able to enter into the narrow space.
To prevent this situation, the longest edge of the triangle formed by the three robots is made longer than the other edges.
To further address this issue, we introduce an acceleration component, $a_{i}^{da}$, that aims to extend the longest edge of a triangle consisting of robot triplets $i, j$, and $k$.
Specifically, if there is an obstacle between either $(i,j) \in \mathcal{E}_m$ or $(j,k) \in \mathcal{E}_m$, and $(i,k) \in \mathcal{E}_m$, and $\|x_{ij}\| > \|x_{kj}\| > d_c + \delta_c, d_c + \delta_c > \|x_{ki}\| > d_c$, then robot $i$ will generate an acceleration to extend the longest edge $(i,j)$ of the triangle further than the other edges.
We denote the set of connected pairs of robots $(j, k)$ satisfying these conditions as $\mathcal{DA}_i$.
To enforce the deadlock avoidance constraint, the acceleration $a_{i}^{da}$ is determined as
\begin{align}
    a_{i}^{da}      & = - \mathrm{\nabla}_{x_i}\Psi_i^{da}(x), \\
    \Psi_i^{da} (x) & = \sum_{(j,k)\in\mathcal{DA}_i} {\phi^{da} (\|x_{ji}\| - \|x_{jk}\|)}, \\
    \phi^{da} (z)   & = \begin{cases}
                            \frac{1}{z},                        & z \geq \frac{1}{\sqrt{\beta_{da}}}, \\
                            -\beta_{da} z + 2\sqrt{\beta_{da}}, & 0 \leq z \leq \frac{1}{\sqrt{\beta_{da}}}.
                        \end{cases}
\end{align}
Here, the corresponding component of $a_{i}^{da}$ will have a large value when the lengths of the two longest links in a robot triplet converge towards equal lengths, thereby averting the isosceles triangle shape.
The constant $\beta_{da} > 0$ is defined to limit the maximum acceleration for deadlock avoidance, as excessively large accelerations can unnecessarily alter robot movements.

To maintain cohesion among the robots, we introduce the following attractive APF $\Psi_i^{ag}(x)$, which promotes cohesion among the robots in the absence of obstacles.
\begin{align}
    \Psi_i^{ag}(x) & = \frac{1}{n_i^{coh}} \sum_{j \in \mathcal{S}_i\setminus \mathcal{N}_i^\sigma} {\phi^{coh}(\|x_{ij}\|)} + \sum_{(j,k)\in\mathcal{CB}_i}{\phi^{cb}(\|\varphi(x_{ji},x_{jk})\|)}, \label{eq:Psi_agg} \\ 
    \phi^{coh}(z)  & = \begin{cases}
                           \frac{1}{2}(z-d_m)^2, & \|z\|>d_m,\ \mathcal{O}_i = \emptyset \\
                           0,                    & \text{otherwise}
                       \end{cases}, \\
    \mathcal{CB}_i & = \{(j,k)\in\mathcal{N}_i^\sigma ~|~ \|x_{jk}\|>d_m,\ \mathcal{O}_i = \emptyset\},\\
    \phi^{cb}(z)   & = \begin{cases}
                           \frac{1}{2}(d_{del}+\delta_{del}-z)^2, & \|z\|< d_{del}+\delta_{del} \\
                           0,                                     & \text{otherwise}
                       \end{cases}.
\end{align}
\normalsize
The $\phi^{coh}$ component of $\Psi_i^{ag}$ motivates robot $i$ to approach the robots within its sensing range, denoted by the set $\mathcal{S}_i$.
This cohesive behavior is achieved by minimizing the distance between robot $i$ and the robots in $\mathcal{S}_i \setminus \mathcal{N}_i^\sigma$, which represents the set of robots that have a direct LOS to robot $i$ but are not included in $\mathcal{N}_i^\sigma$.
The cardinality of $\mathcal{S}_i\setminus\mathcal{N}_i^\sigma$ denoted as $n_i^{coh}$. 
On the other hand, the $\phi^{cb}$ component of $\Psi_i^{ag}$ disrupts the straight-line formation between robot $i$ and its neighboring robots, facilitating the quick creation of links to them and establishing new LOS for them.
The $\phi^{cb}$ component is defined based on the distance between robot $i$ and the nearest point on the line $\mathcal{L}(x_j,x_k)$, where $d_{del}+\delta_{del}$ represents the intended distance margin for establishing new LOS.
The $\mathcal{CB}_i$ set is used to determine the $\phi^{cb}$ component of the APF, encompassing the pair of robots linked with robot $i$ and separated by a distance greater than the maximum link distance $d_m$.
The acceleration to enforce aggregation is determined as follows, where $\beta_{ag}$ is a design constant,
\begin{align}
    a_{i}^{ag} = -\beta_{ag}\mathrm{\nabla}_{x_i}\Psi_i^{ag}(x).
\end{align}
Overall, the APF $\Psi_i^{ag}(x)$ enables the follower robots to gather around the leader robot, thus preventing the swarm from dividing and allowing for better coordination among all robots.

\subsection{Verification of constraint enforcement (Step 3)}
For the desired input $a_i$ determined in Step 2, the control input $u_i$ is calculated by solving a numerical optimization problem considering the given CBF constraints.
Since multiple constraints are severely imposed in our control problem, the control input satisfying all the constraints does not necessarily exist.
In this paper, a classical approach utilizing soft bounds \cite{Maciejowski2002} is adopted to ensure the feasibility of the optimization problem.

To describe the optimal control problem in a compact form, we describe the CBF-based constraints in Section \ref{sec:apf_cbf} as follows:
\begin{align}
    A u_i + b \geq 0, \label{eq:const_compact}
\end{align}
where the inequality of each row is one of the constraints in \eqref{eq:Gc_bar_distributed}, \eqref{eq:Gm_bar_distributed}, \eqref{eq:Go_bar}, and \eqref{eq:Gls_bar_distributed}.

Robot $i $ $(=1,\cdots, N)$ determines $u_i$ by solving the following optimization problem at each time step.
\begin{align}
    \begin{aligned}
         & \min_{u_i \in \mathbb{R}^n,\ \epsilon \in \mathbb{R}}{ \{\|u_i - a_i\|^2 + \rho \epsilon \}} \\
         & \text{subject to } \\
         & \quad A  u_i + b + \epsilon \mathbf{1} \geq 0, \\
         & \quad \|u_i \| \leq \eta, \\
         & \quad \epsilon \geq 0. \label{eq:G_SDP}
    \end{aligned}
\end{align}
where $\mathbf{1}$ is the vector with all entries being one, and $\epsilon$ is a slack variable which is non-zero only if the constraint in \eqref{eq:const_compact} is violated.
By heavily penalizing $\epsilon$ in the cost function using a large weight $\rho > 0$, the optimal solution to the problem in \eqref{eq:G_SDP} tries to minimize constraint violations as much as possible.
This optimization is reduced to a semi-definite programming problem and thus can be solved efficiently.

\subsection{Recovery control} \label{sec:recoverry}
In Step 3 of the algorithm, we relax the CBF-based constraints to obtain a control input.
This relaxation may lead to violations of the original position-based constraints defined in Section \ref{sec:problem}.
To address this issue, a recovery control law is introduced to manage constraint violations and return the robot to a safe zone.
The recovery control law is activated when one or more constraints are violated.
In such cases, the control input for the robot is determined solely by computing acceleration for the violated constraints, disregarding the other acceleration components.
More specifically, the recovery control input is computed as follows,
\begin{align}
    u_i         &= a_i^r - k_{r} v_i , \label{eq:recovery_control} \\
    a_i^r       &= c_m a_{i}^{mr} + c_c a_{i}^{cr} + c_{ob} a_{i}^{obr} + c_{ls} a_{i}^{lsr}, \\
    a_{i}^{mr}  &= \sum_{j\in\mathcal{R}_i^m}{\frac{\|x_{ji}\| - d_m}{\bar{d}_m - d_m}} {\frac{x_{ji}}{\|x_{ji}\|}},\\
    a_{i}^{cr}  &= \sum_{j\in\mathcal{R}_i^c}{\frac{\|x_{ji}\| - d_c}{d_c - \underline{d}_c}} {\frac{x_{ji}}{\|x_{ji}\|}},\\
    a_{i}^{obr} &= \sum_{x_o\in\mathcal{R}_i^{ob}}{\frac{\|x_{oi}\| - d_o}{d_o - \underline{d}_o}} {\frac{x_{oi}}{\|x_{oi}\|}},\\
    a_{i}^{lsr} &= \sum_{(x_o, j)\in\mathcal{R}_i^{ls}}{\frac{\|x_{oq}\| - d_{ls}}{d_{ls} - \underline{d}_{ls}}} {\frac{x_{oq}}{\|x_{oq}\|}},\\
    x_{oq}      &= -\varphi(x_{oi},x_{ji}),
\end{align}
where $c_m > 0, c_c > 0, c_{ob} > 0, c_{ls} > 0$ are weights assigned for each recovery acceleration component.
Here, $\mathcal{R}_i^m$ denotes the robots that violate condition \eqref{eq:c1}, while $\mathcal{R}_i^c$ refers to robots in violation of condition \eqref{eq:c2}. The set $\mathcal{R}_i^{ob}$ includes obstacle points that breach condition \eqref{eq:c3}, and $\mathcal{R}_i^{ls}$ comprises both robots and the corresponding obstacle points that violate the LOS constraint specified in \eqref{eq:c4}.

\section{Approximate method without numerical optimization} \label{sec:appr}
Although the method in Section \ref{sec:ctrl_alg} requires to solve a numerical optimization problem at each time step in Step 3, robots may not always possess sufficient computational capacity for this task in some cases.
Thus, many control methods for multi-agent systems adopt simple control algorithms without solving numerical optimization \cite{Sakai2018, Mondal2018, Nomura2021, Qiao2022, Olfati2006, Su2009, Sakai2017}.
These methods determine control input by employing a combination of repulsive and attractive acceleration components based on APFs.
In other words, high priority is given to constraints with large repulsive acceleration, while constraints with small repulsive acceleration are ignored.
Similarly, the proposed approximate method determines the direction of the control input using $a_i$ in \eqref{eq:a_des}, which is the sum of repulsive and attractive acceleration components.
Only the magnitude of the control input is adjusted, as follows:
\begin{align}
    u_i = \lambda \frac{a_i}{\| a_i \|} ,~~~ \lambda \in [0, \eta]. \label{eq:u_appr}
\end{align}

For the $r$th row of the constraint in \eqref{eq:const_compact}, $A_r u_i + b_r \geq 0$, we determine an allowable region $[\lambda_r^{min}, \lambda_r^{max}]$ depending on situations as follows:
\begin{enumerate}[label=, leftmargin=0pt, labelsep=0pt]
    \item Case 1: $ A_r \frac{a_i}{\|a_i \|} \geq 0 $ and $ b_r \geq 0 $.\\
          In this case, $\lambda$ can take any positive value as the condition will always be satisfied.
          Thus, the limits are set as $\lambda_r^{min}=0, \lambda_r^{max}=\eta$.

    \item Case 2: $ A_r \frac{a_i}{\|a_i \|} < 0 $ and $ b_r \geq 0 $.\\
          In this case, the constraint will be satisfied for $\lambda \leq -\frac{b_r \|a_i \|}{A_r a_i}$.
          Thus, the limits are set as $\lambda_r^{min}=0, \lambda_r^{max}=-\frac{b_r \|a_i \|}{A_r a_i}$.

    \item Case 3: $ A_r \frac{a_i}{\|a_i \|} \leq 0 $ and $ b_r < 0 $.\\
          In this scenario, the inequality cannot be satisfied for a nonnegative $\lambda$.
          Since there is no choice other than to ignore this constraint, the limits are set as $\lambda_r^{min}=0, \lambda_r^{max}=\eta$.

    \item Case 4: $ A_r \frac{a_i}{\|a_i \|} > 0 $ and $ b_r < 0 $.\\
          In this case, the constraint will be satisfied for $\lambda \geq -\frac{b_r \|a_i \|}{A_r a_i}$.
          Thus, if $\eta < -\frac{b_r \|a_i \|}{A_r a_i}$, the limits are set as $\lambda_r^{min}=0, \lambda_r^{max}=\eta$, since this constraint cannot be satisfied for $\lambda \in [0, \eta]$.
          Otherwise, the limits are set as $\lambda_r^{min}=-\frac{b_r \|a_i \|}{A_r a_i}, \lambda_r^{max}=\eta$.
\end{enumerate}

To determine $\lambda$ in \eqref{eq:u_appr}, we use the following $\lambda^{min}$ and $\lambda^{max}$
\begin{align}
    \lambda^{min} & := \max_r \lambda^{min}_r, \\
    \lambda^{max} & := \min_r \lambda^{max}_r.
\end{align}
If $\lambda^{min} \leq \lambda^{max}$, it is easy to determine $\lambda$ depending on the $\|a_i\|$.
On the other hand, $\lambda^{max} < \lambda^{min}$ implies that constraints in Cases 2 and 4 conflict with each other.
In such cases, we ignore $\lambda^{min}$ for the following reason:
The value of $\lambda^{min}$ is typically large when the difference between the direction of $a_i$ and a repulsive acceleration of constraints in Case 4 is large.
Since we prioritize constraints with repulsive forces in directions similar to $a_i$, the priority of constraints in Case 4 with a large $\lambda^{min}$ is low.
Therefore, we ignore $\lambda^{min}$ if $\lambda^{max} < \lambda^{min}$.
In summary, we determine $\lambda$ as follows:
\begin{align}
    \lambda & = \begin{cases}
                    \lambda^{max}, & \|a_i\| \geq \lambda^{max}                 \\
                    \lambda^{min}, & \|a_i\| < \lambda^{min} \leq \lambda^{max} \\
                    \| a_i \|,     & \text{otherwise.}
                \end{cases}
\end{align}

\section{Simulations} \label{sec:sim}
We applied the proposed method to a quadrotor model given in \cite{Corke2017} with inner-loop PD controllers for attitude and velocity. 
The simulations were run on a computer with an Intel Core i9 12900k processor, 32 GB of RAM, and an NVIDIA GeForce RTX 3090 graphics card.
The control input was computed at intervals of $\Delta t = 0.1$ s, and the system parameters were set as listed in Table \ref{table:sim_setting}.
The control input $u_i$ determined by the proposed method was passed as the target velocity $v_i(t + \Delta t) = v_i(t) + \Delta t u_i(t)$ to the quadrotor model.
Furthermore, considering the maximum acceleration limit, we set $\mu_m = \mu_c = \mu_{ob} = \mu_{ls} = \eta$.
Additionally, we limited the maximum speed of the leader to $0.1\ m/s$, enabling the follower robots to aggregate around the leader in obstacle-free environments.

\begin{table}
    \renewcommand{\arraystretch}{1.5}
    \tbl{Parameters used in the simulation.}
    {\begin{tabular}{|c|c|c|c|c|c|c|}\hline
        \rowcolor[rgb]{0.9, 0.9, 0.9}
        $d_s$                & $\bar{d}_{m}$ & $d_m$     & $\underline{d}_{c}$ & $d_c$          & $\underline{d}_{o}$ & $d_o$        \\ \hline
        2                    & 1.9           & 1         & 0                   & 0.1            & 0                   & 0.1          \\ \hline \hline
        \rowcolor[rgb]{0.9, 0.9, 0.9}
        $\underline{d}_{ls}$ & $d_{ls}$      & $d_{del}$ & $\delta_m$          & $\delta_{del}$ & $\alpha_m$          & $\alpha_c$   \\ \hline
        0                    & 0.05          & 0.05      & 0.05                & 0.05           & 0.1                 & 0.1          \\ \hline \hline
        \rowcolor[rgb]{0.9, 0.9, 0.9}
        $\alpha_{ob}$        & $\alpha_{ls}$ & $\beta_m$ & $\beta_c$           & $\beta_{ob}$   & $\beta_{ls}$        & $\beta_{da}$ \\ \hline
        0.4                  & 0.2           & $0.01$    & $0.01$              & $0.001$        & $0.001$             & $0.01$       \\ \hline \hline
        \rowcolor[rgb]{0.9, 0.9, 0.9}
        $\beta_{ag}$         & $\rho$        & $c_m$     & $c_c$               & $c_o$          & $c_{ls}$            & $\eta$       \\ \hline
        $0.5$                & $10^6$        & 1         & 1                   & 0.4            & 0.5                 & 1            \\ \hline
    \end{tabular}}
    \label{table:sim_setting}
    \renewcommand{\arraystretch}{1}
\end{table}

\begin{figure}[h]
    \centering
    \includegraphics[width=0.8\linewidth]{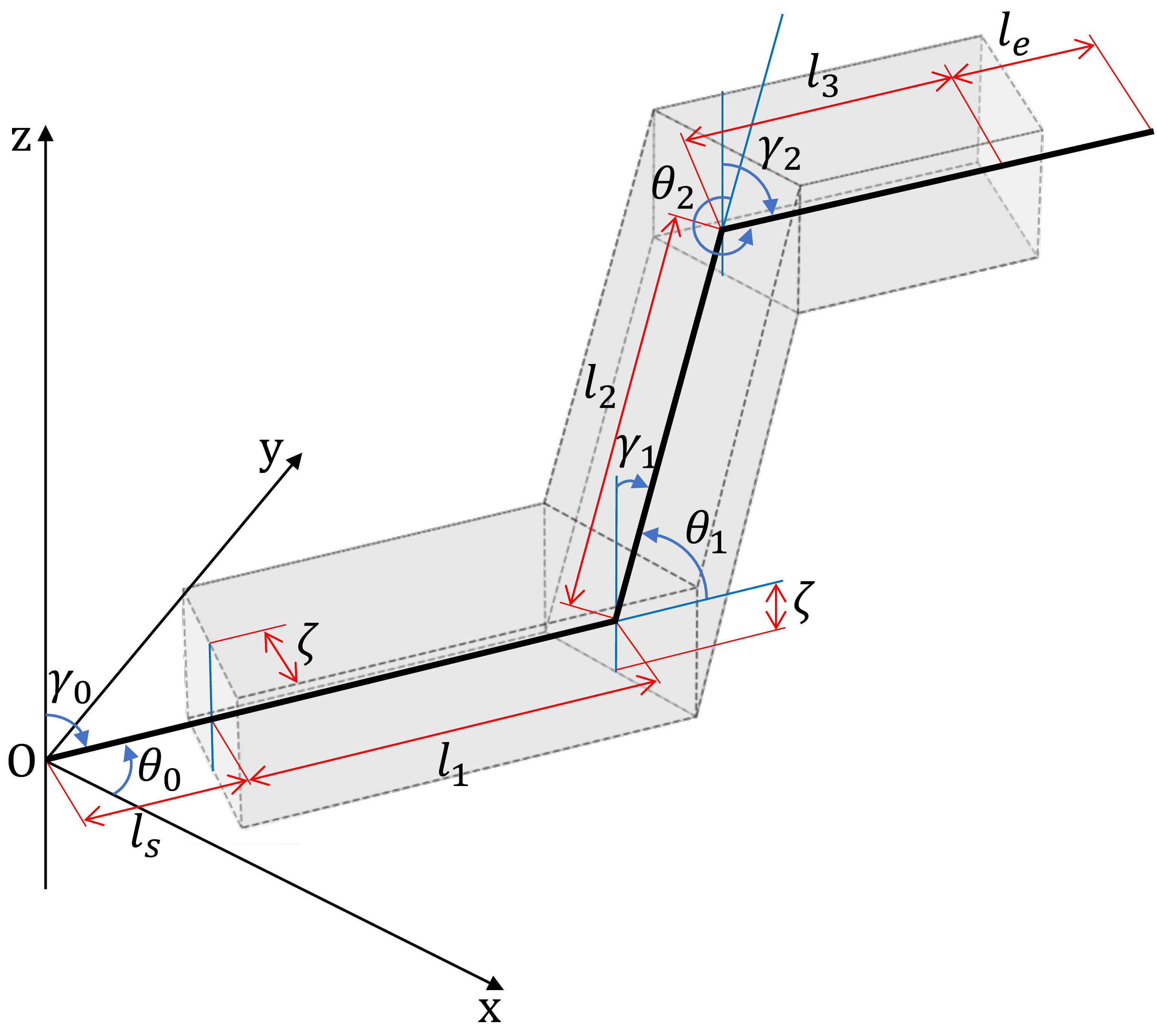}
    \caption{Square tunnel obstacle configuration.}
    \label{fig:Tunnel_obs}
\end{figure}

We used the obstacle-laden environment configuration depicted in Figure \ref{fig:Tunnel_obs}.
The leader is given a target path that passes through the center of the tunnel, starting from the origin, as illustrated by the thick solid line.
The obstacle-laden environment is designed as a polyhedron tunnel, with plane obstacles positioned on all four sides of the path (left, right, top, bottom).
To define the 3D relative angle between the $(n+1)$th line segment and the $n$th line segment, we use a pair of angles, denoted as $(\theta_n \neq 0, \gamma_n  \neq 0)$ for $n = 1, \dots, M+1$.
Here, $\theta_n$ represents the azimuth angle of the $(n+1)$th line segment with respect to the $n$th line segment, and $\gamma_n$ represents the elevation angle of the $(n+1)$th line segment with respect to the plane perpendicular to the $n$th line segment.
Thus, $M$ represents the number of corners in the target path.
The first and last line segments were divided into parts inside and outside the obstacle area, with $l_s$ and $l_e$ representing the lengths of the parts from the origin to the center of the first obstacle segment and from the center of the last segment to the end of the target path, respectively.
Additionally, the length of the $n$th line segment inside the obstacle area is defined as $l_n$.
The minimum distance between an obstacle and the leader's target path is denoted as $\zeta$.

First, simulations were conducted for a single-segment tunnel obstacle ($M=0$) with various path widths ($2\zeta$).
Other parameters were set as $\theta_0 = 0, \gamma_0 = \pi/2, l_s = 0.5, l_1 = (N-1)d_m/2$, and $l_e = 20$.
For each path-width, a total of 25 sets of simulation trials were conducted by placing $N=10$ robots in different initial positions.
Since the allowable minimum distance between a robot and an obstacle is $d_o = 0.1$ m, a $\zeta$ value below 0.1 m was not feasible.
Therefore, we performed simulations for cases where $\zeta$ ranged from 0.105 m to 0.4 m. 

\begin{figure}
    \centering
    \includegraphics[width=0.95\linewidth]{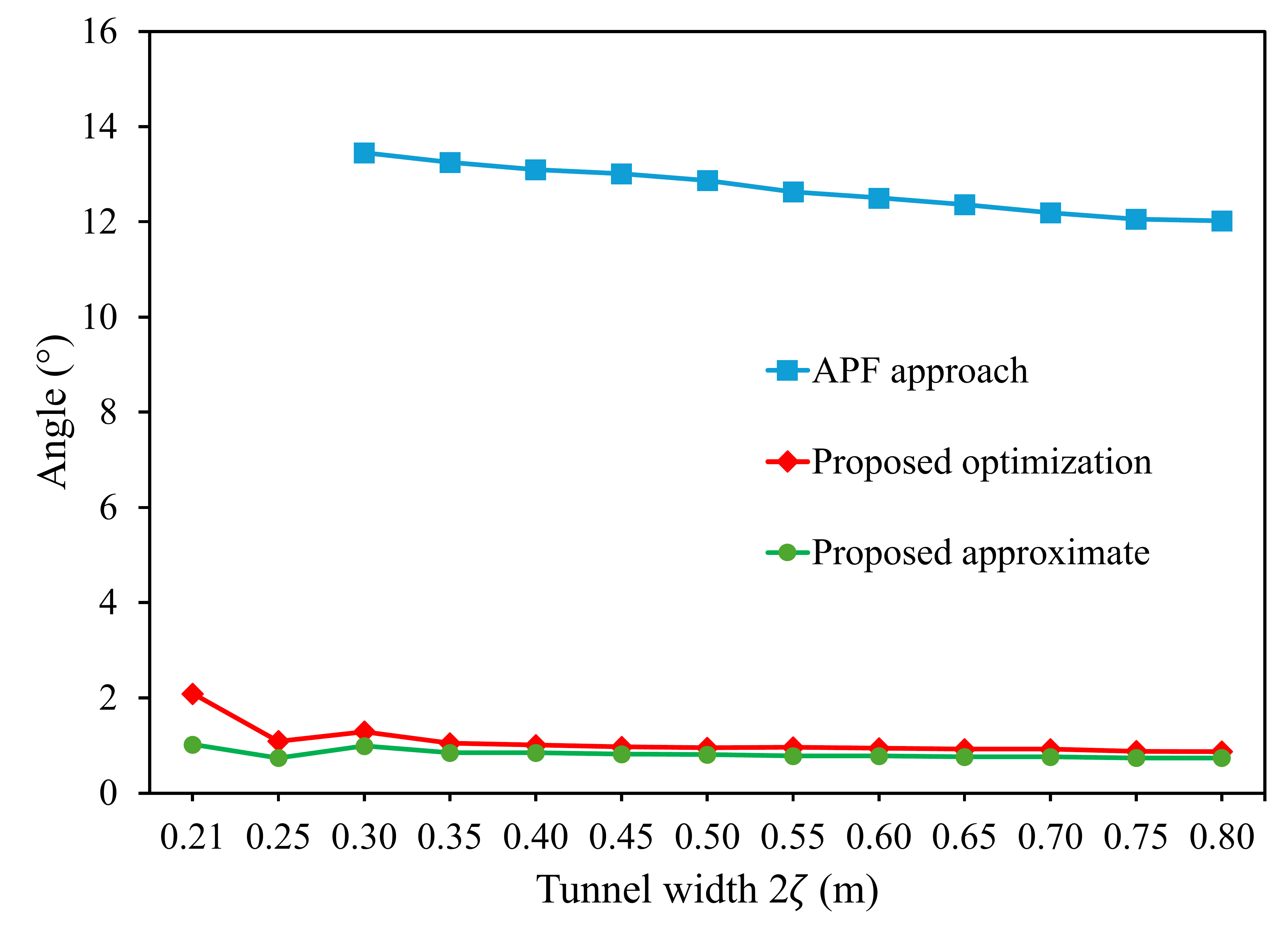}
    \caption{Mean angle differences between consecutive target velocities for different path widths.}
    \label{fig:vibration_analysis}
\end{figure}

To analyze the oscillatory movement, the angle difference between control inputs at consecutive time steps of the follower robots was calculated.
The solid lines in Figure \ref{fig:vibration_analysis} represent the mean angle difference for each path width of the tunnel, with data collected from simulation trials.
The analysis was conducted from the start of the simulation until the last robot passed through the tunnel.
The results of the APF method for $2\zeta = 0.21$ and 0.25 are not shown, since the robots were unable to pass through the tunnel successfully.
This failure occurred for either of two reasons: the robots got stuck in the tunnel or some of them were left behind since the connectivity of the sensing network $\mathcal{G}_s$ was lost.
The proposed approaches exhibit a low mean, indicating consistent direction changes in consecutive control inputs and thus minimal vibratory behavior in the robots.
In contrast, the APF-based approach shows a high mean, indicating inconsistent direction changes and significant oscillatory movement of the robots.

\begin{figure*}
   \centering
   \subfloat[]{\includegraphics[width=0.5\linewidth]{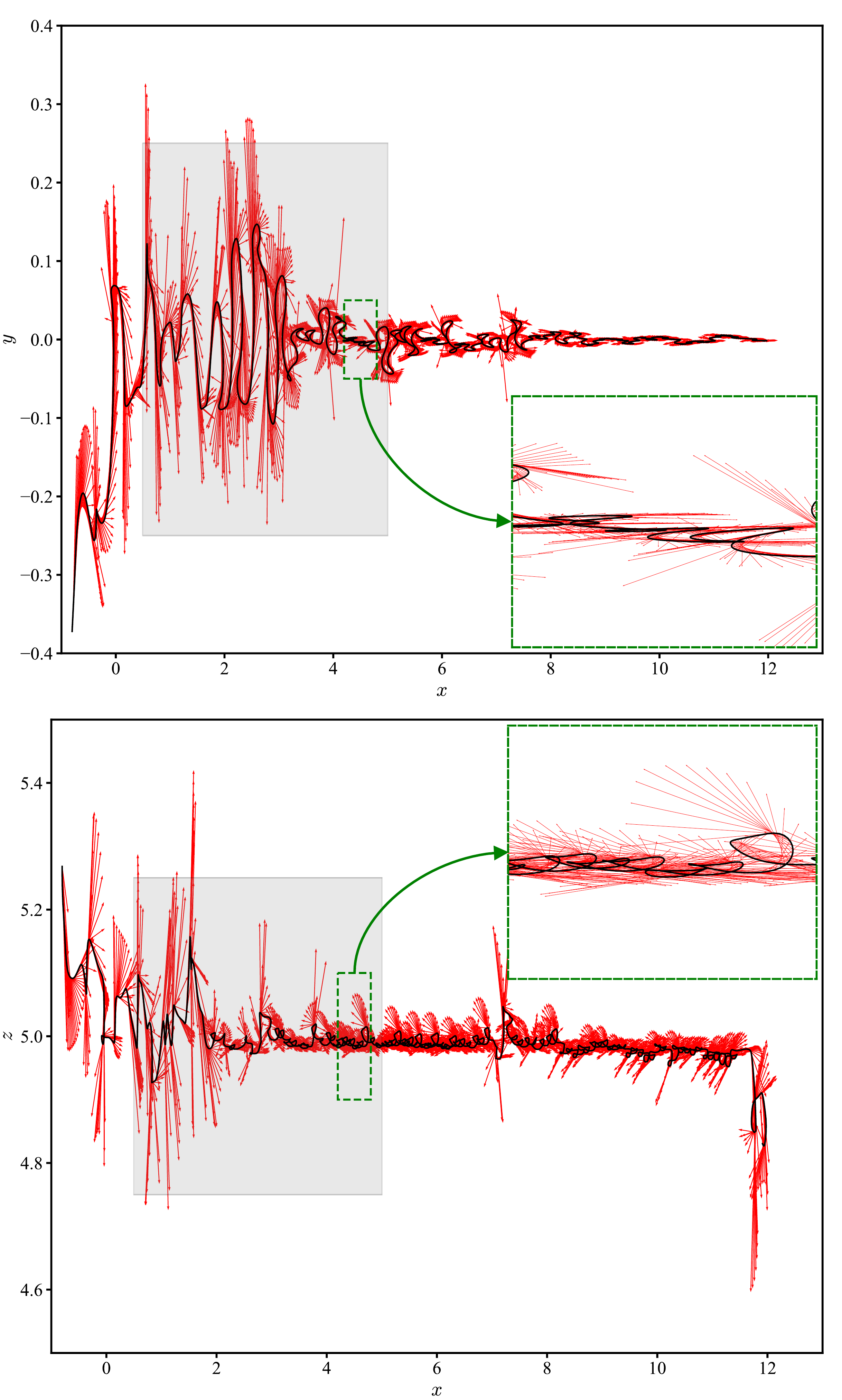}}
   \subfloat[]{\includegraphics[width=0.5\linewidth]{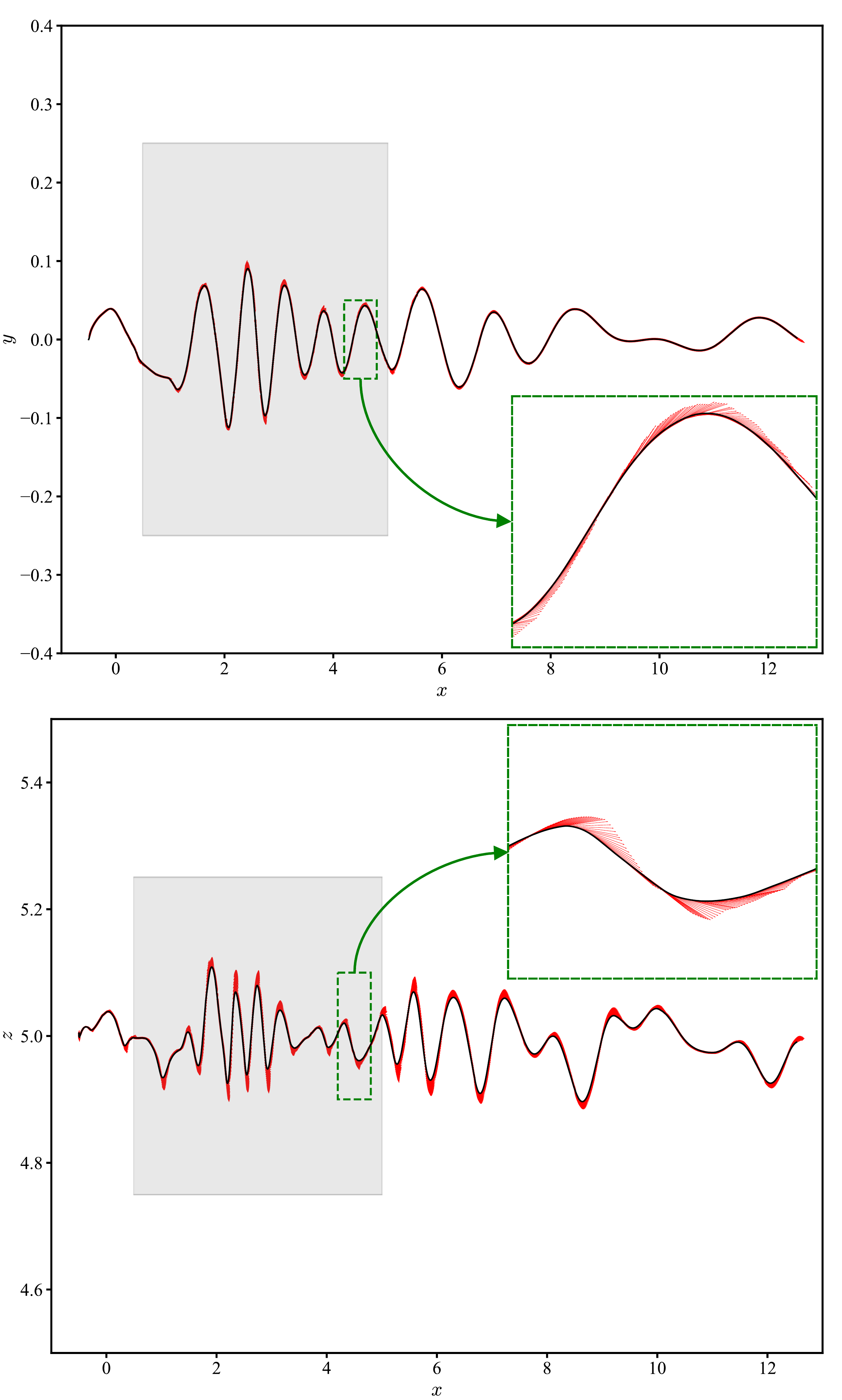}}
   \caption{Executed trajectory of the first follower robot: (a) APF approach, (b) Proposed approximate method.
   The red arrows indicate the direction of target velocity commands sent to the quadrotor model, while the black line represents the executed trajectory.
   The top row shows the trajectories in the XY-plane, and the bottom row shows the trajectories in the XZ-plane.
   The obstacle is represented by the gray-shaded area.}
   \label{fig:CBF_vs_APF_trajectory}
\end{figure*}

A typical example of responses that indicate differences in oscillation is illustrated in Figure \ref{fig:CBF_vs_APF_trajectory} for the first follower robot.
The oscillatory behavior of the robots in the APF approach is clear, as the target velocity commands are arbitrary and inconsistent.
In contrast, the proposed approach demonstrates a steady and minimal change in the target velocity command.

Table \ref{table:Rate_of_constraint_violations} compares constraint violations, while Table \ref{table:CPU_time} shows the mean computation times for Step 3 between the optimization-based and approximate methods.
To assess the constraint violations, we tallied for each robot the time steps in which any of the constraints in \eqref{eq:c1}--\eqref{eq:c4} were violated, across all 25 simulation trials.
We then calculated the percentage of these violated time steps relative to the total number of time steps for each tunnel widths.
As expected, the approximate method shows a slightly higher rate of constraint violations compared to the optimization-based method, albeit with much shorter computation times, as illustrated in Table \ref{table:CPU_time}.

\renewcommand{\arraystretch}{1.5}
\begin{table}
    \tbl{Rate of constraint violation in percentage (\%) for single segment tunnel obstacle.}
    {\begin{tabular}{|c|c|c|c|} \hline
        \rowcolor[rgb]{0.9, 0.9, 0.9}
        Tunnel width ($2\zeta$) & APF approach  & Proposed optimization & Proposed approximate  \\ \hline
        0.21                    & -             & 2.2$\times 10^{-3}$   & 2.7$\times 10^{-2}$   \\ \hline
        0.25                    & -             & 5.9$\times 10^{-4}$   & 1.3$\times 10^{-2}$   \\ \hline
        0.30                    & 4.2           & 0                     & 7.4$\times 10^{-4}$   \\ \hline
        0.35                    & 2.9           & 0                     & 4.3$\times 10^{-4}$   \\ \hline
        0.40                    & 2.0           & 0                     & 6.0$\times 10^{-4}$   \\ \hline
        0.45                    & 1.8           & 0                     & 1.3$\times 10^{-4}$   \\ \hline
        0.50                    & 1.7           & 0                     & 9.4$\times 10^{-4}$   \\ \hline
        0.55                    & 1.6           & 0                     & 4.3$\times 10^{-4}$   \\ \hline
        0.60                    & 1.6           & 0                     & 0                     \\ \hline
        0.65                    & 1.6           & 0                     & 0                     \\ \hline
        0.70                    & 1.6           & 0                     & 6.0$\times 10^{-4}$   \\ \hline
        0.75                    & 1.6           & 0                     & 0                     \\ \hline
        0.80                    & 1.6           & 0                     & 0                     \\ \hline
    \end{tabular}}
    \label{table:Rate_of_constraint_violations}
\end{table}
\renewcommand{\arraystretch}{1}

\renewcommand{\arraystretch}{1.5}
\begin{table}
    \tbl{Mean CPU time in milliseconds for step 3 in single segment tunnel obstacle for proposed methods.}
    {\begin{tabular}{|c|c|c|} \hline
        \rowcolor[rgb]{0.9, 0.9, 0.9}
        Tunnel width ($2\zeta$) & Optimization  & Approximate   \\ \hline
        0.21                    & 23.75         & 0.07          \\ \hline
        0.25                    & 29.78         & 0.06          \\ \hline
        0.30                    & 25.42         & 0.07          \\ \hline
        0.35                    & 31.75         & 0.06          \\ \hline
        0.40                    & 27.51         & 0.07          \\ \hline
        0.45                    & 33.50         & 0.06          \\ \hline
        0.50                    & 29.05         & 0.08          \\ \hline
        0.55                    & 32.85         & 0.06          \\ \hline
        0.60                    & 29.51         & 0.08          \\ \hline
        0.65                    & 32.41         & 0.06          \\ \hline
        0.70                    & 29.68         & 0.08          \\ \hline
        0.75                    & 32.30         & 0.06          \\ \hline
        0.80                    & 29.54         & 0.08          \\ \hline
    \end{tabular}}
    \label{table:CPU_time}
\end{table}
\renewcommand{\arraystretch}{1}

Next, we conducted simulations for more complex configurations listed in Table \ref{table:obstacle_config}.
Table \ref{table:CBF_extensive_violation} presents the constraint violation rates across 25 simulation trials for obstacle configurations 1-3, while Table \ref{table:CBF_extensive_CPU_time} shows the CPU time taken in step 3.
The results for $2\zeta \geq 0.4$ are shown, since the robots fail to navigate through the tunnel successfully in some cases for $2\zeta \leq 0.3$.
These results show that both proposed methods exhibit few constraint violations.
However, the computation time for step 3 is very short for the approximate method, around 0.08 milliseconds, while the optimization-based method averages around 30.94 milliseconds.
Therefore, the proposed approximate method is expected to be effective in situations where the computer is not fast enough for the optimization-based method to be applied.

\renewcommand{\arraystretch}{2}
\begin{table}[!h]
    \tbl{Complex obstacle configurations.}
    {\begin{tabular}{|c|l|l|l|} \hline
        \rowcolor[rgb]{0.9, 0.9, 0.9}
        Config.  & \multicolumn{1}{c|}{$l_1, \dots$}                                & \multicolumn{1}{c|}{$\theta_1, \dots$} & \multicolumn{1}{c|}{$\gamma_1, \dots$} \\ \hline
        1. $M=1$ & $\frac{(N-1)d_m}{2}$, $\frac{(N-1)d_m}{4}$                       & $\frac{\pi}{3}$                        & $\frac{\pi}{2}$                        \\ \hline
        2. $M=1$ & $\frac{(N-1)d_m}{2}$, $\frac{(N-1)d_m}{4}$                       & $\frac{\pi}{2}$                        & $\frac{\pi}{2}$                        \\ \hline
        3. $M=2$ & $\frac{(N-1)d_m}{2}$, $\frac{(N-1)d_m}{4}$, $\frac{(N-1)d_m}{2}$ & $\frac{\pi}{2}$, $-\frac{\pi}{2}$      & $\frac{\pi}{2}$, $\frac{\pi}{2}$       \\ \hline
    \end{tabular}}
    \label{table:obstacle_config}
\end{table}
\renewcommand{\arraystretch}{1}

\renewcommand{\arraystretch}{1.5}
\begin{table*}[!h]
    \tbl{Rate of constraint violation in percentage (\%) for Config. 1-3.}
    {\begin{tabular}{|c|c|c|c|c|c|c|} \hline
        \rowcolor[rgb]{0.9, 0.9, 0.9}
        Tunnel              & \multicolumn{2}{c|}{Config. 1}                & \multicolumn{2}{c|}{Config. 2}                & \multicolumn{2}{c|}{Config. 3}                \\ \cline{2-7}
        \rowcolor[rgb]{0.9, 0.9, 0.9}
        width ($2\zeta$)    & Optimization          & Approximate           & Optimization          & Approximate           & Optimization          & Approximate           \\ \hline
        0.4                 & 0	                    & 1.0$\times 10^{-2}$	& 0	                    & 1.7$\times 10^{-2}$	& 1.3$\times 10^{-4}$	& 9.6$\times 10^{-3}$   \\ \hline
        0.5                 & 0	                    & 9.9$\times 10^{-4}$	& 0	                    & 2.6$\times 10^{-3}$	& 0	                    & 2.5$\times 10^{-3}$   \\ \hline
        0.6                 & 1.3$\times 10^{-3}$   & 0	                    & 5.0$\times 10^{-3}$	& 0	                    & 3.6$\times 10^{-3}$	& 0                     \\ \hline
        0.7                 & 5.3$\times 10^{-4}$   & 0	                    & 4.2$\times 10^{-3}$	& 1.1$\times 10^{-4}$	& 3.5$\times 10^{-3}$	& 1.6$\times 10^{-4}$   \\ \hline
        0.8                 & 0	                    & 0	                    & 3.7$\times 10^{-3}$	& 1.1$\times 10^{-4}$	& 3.2$\times 10^{-3}$	& 0                     \\ \hline
    \end{tabular}}
    \label{table:CBF_extensive_violation}
\end{table*}
\renewcommand{\arraystretch}{1}

\renewcommand{\arraystretch}{1.5}
\begin{table*}[!h]
    \tbl{Mean CPU time in milliseconds for Config. 1-3.}
    {\begin{tabular}{|c|c|c|c|c|c|c|} \hline
        \rowcolor[rgb]{0.9, 0.9, 0.9}
        Tunnel              & \multicolumn{2}{c|}{Config. 1}   & \multicolumn{2}{c|}{Config. 2}    & \multicolumn{2}{c|}{Config. 3}    \\ \cline{2-7}
        \rowcolor[rgb]{0.9, 0.9, 0.9}
        width ($2\zeta$)    & Optimization & Approximate       & Optimization  & Approximate       & Optimization & Approximate        \\ \hline
        0.4                 & 27.69        & 0.07              & 31.42         & 0.07              & 32.09        & 0.07               \\ \hline
        0.5                 & 28.84        & 0.07              & 31.05         & 0.08              & 32.61        & 0.08               \\ \hline
        0.6                 & 29.87        & 0.08              & 31.53         & 0.08              & 31.67        & 0.08               \\ \hline
        0.7                 & 30.09        & 0.08              & 31.49         & 0.09              & 31.67        & 0.08               \\ \hline
        0.8                 & 30.74        & 0.08              & 31.62         & 0.09              & 31.73        & 0.08               \\ \hline
    \end{tabular}}
    \label{table:CBF_extensive_CPU_time}
\end{table*}
\renewcommand{\arraystretch}{1}

\begin{figure*}
   \centering
   \includegraphics[width=0.98\linewidth]{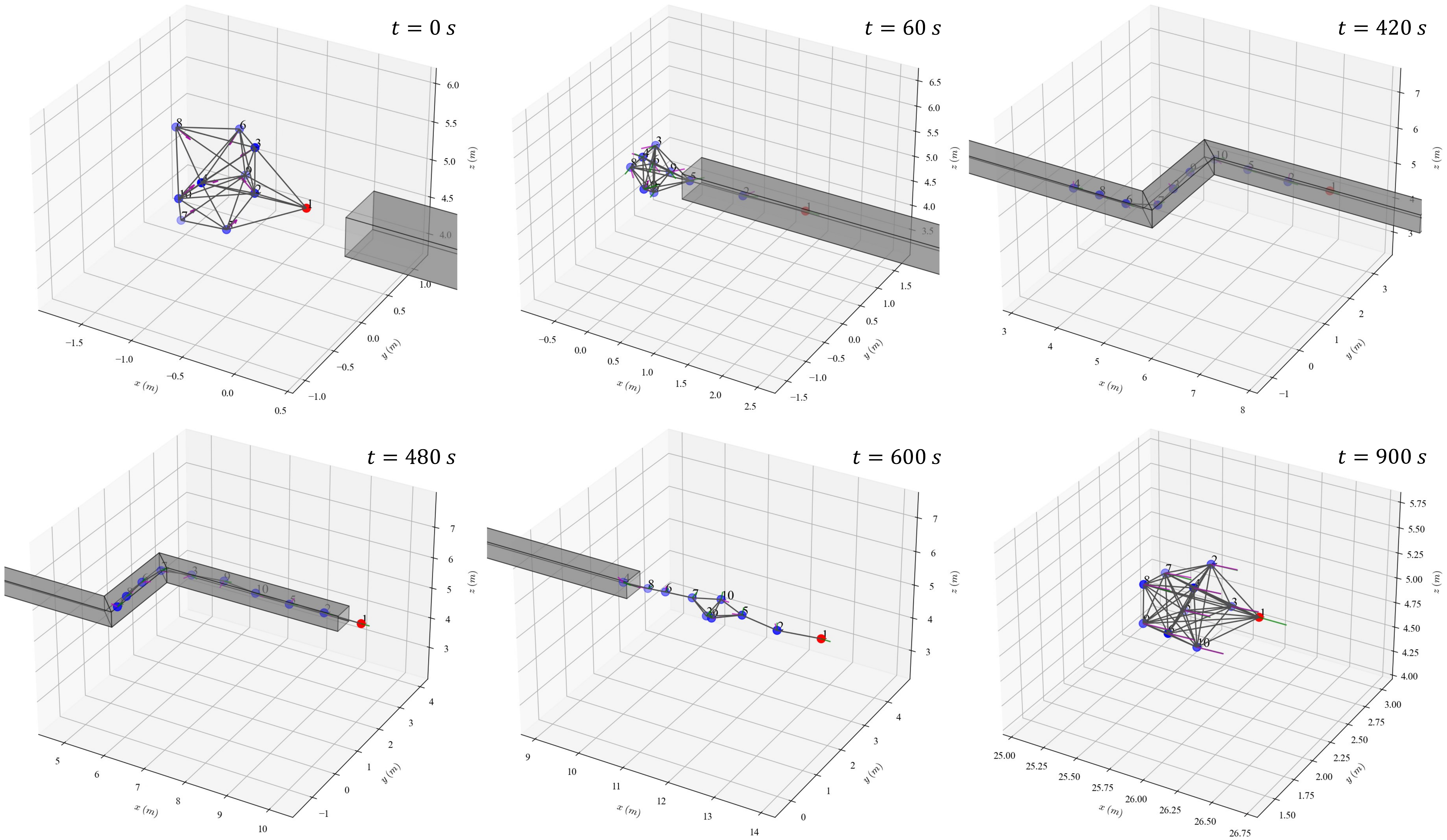}
   \caption{Simulation snapshots of a random trial for the proposed approximate method in obstacle config. 3 with $2\zeta = 0.5$ m.
   The leader and follower robots are represented by red and blue circles, respectively, while the black line segment represents the LOS connectivity between the robots in $\mathcal{G}_\sigma$.
   }
   \label{fig:simulation_snaps}
\end{figure*}

Figure \ref{fig:simulation_snaps} displays snapshots of the proposed approximate method simulation representing a random trial with $2\zeta = 0.5$ m.
The snapshots show the successful disconnection of redundant links by the robots upon entering the tunnel, resulting in a chain formation.
They also demonstrate the proposed algorithm's ability to allow the robots to navigate sharp turns without losing LOS connectivity while traversing critical corners.
Once the robots have passed through the obstacle environment, they effectively aggregate around the leader robot.

\section{Experiment} \label{sec:experiment}
The experiments were conducted in an environment where an obstacle wall was situated in the middle of the leader's target path, featuring a 0.6 m hole in the middle.
The positions of the quadrotors were tracked using the OptiTrack Prime17W motion capture system.
A formation of five Crazyflie 2.1 quadrotors was employed for experimental validation.
Control inputs were computed on a single computer using the proposed method and transmitted to the quadrotors as target velocities with a constant yaw angle.
Note that control inputs were calculated in a decentralized manner for each quadrotor based on local information within its sensing range.
The wall obstacle is approximated using multiple spheres in the same way as in \cite{Nomura2021}.
The obstacle position data were provided to the robots only when they were within sensing range, as obstacle recognition was not the focus of this experiment.
Velocity commands were sent to the robots at every $\Delta t = 0.1$ s, with the same parameters detailed in Table \ref{table:sim_setting} except for $d_s = 1.5, \bar{d}_{m} = 1.45, d_c = 0.2$, and $d_o = 0.15$.

To deal with the downwash created by the quadrotors, the following constraint is introduced to prevent robots from aligning vertically.
\begin{align}
    \|\bar{x}_i(t) - \bar{x}_j(t)\| \geq d_{dw} ,\  \forall j \in \mathcal{N}_i(t), \label{eq:c5}
\end{align}
where $\bar{x}_i$ is defined by making the third element of $x_i$ equal to zero to compute the horizontal distance between robots, and $d_{dw}$ is the minimum allowable distance between robots to prevent the downwash effect.
Similar constraints to \eqref{eq:Gc_bar} were derived using a derivative condition as in \eqref{eq:Gc}.
The parameter values were set as $d_{dw} = 0.2, \alpha_{dw} = 0.5$, and $\beta_{dw} = 10^{-4}$.

\begin{figure}
   \centering
   \includegraphics[width=0.75\linewidth]{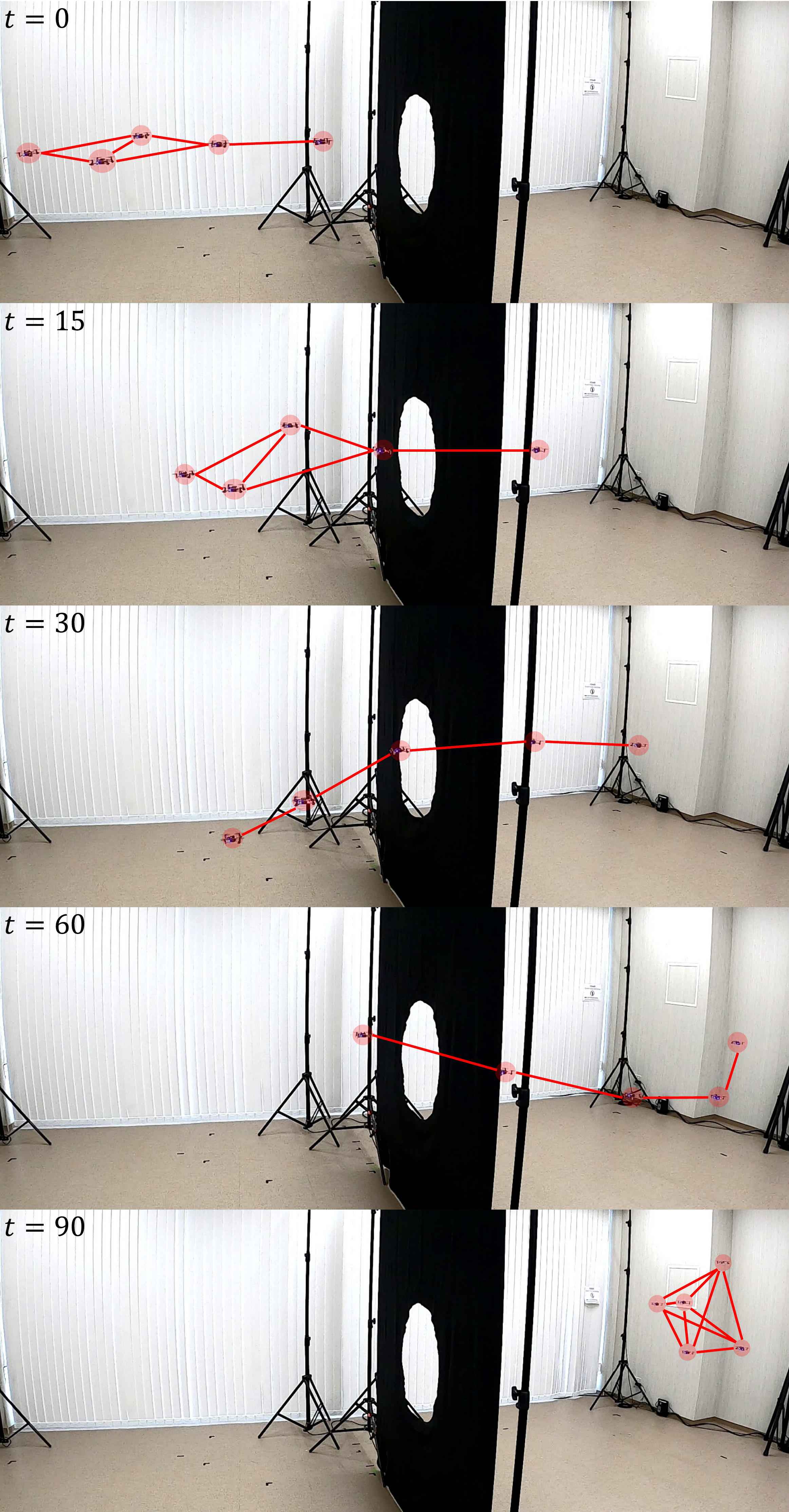}
   \caption{Experiment snapshots of the proposed approximate method.
   The red lines denote the edges of the graph $\mathcal{G}_\sigma$.}
   \label{fig:exp_snaps}
\end{figure}

\begin{figure*}
   \centering
   \subfloat[]{\includegraphics[width=0.33\linewidth]{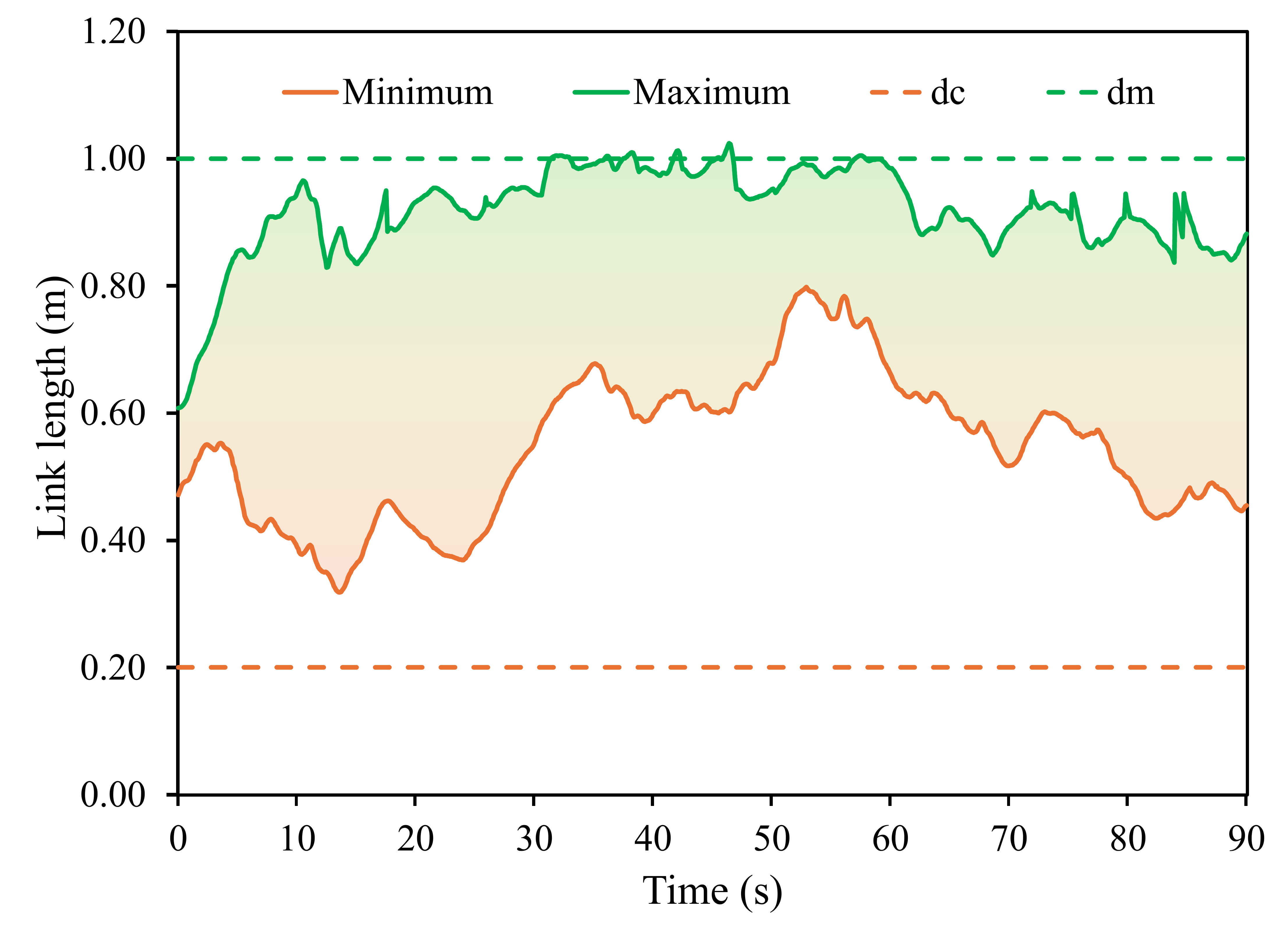}}
   \subfloat[]{\includegraphics[width=0.33\linewidth]{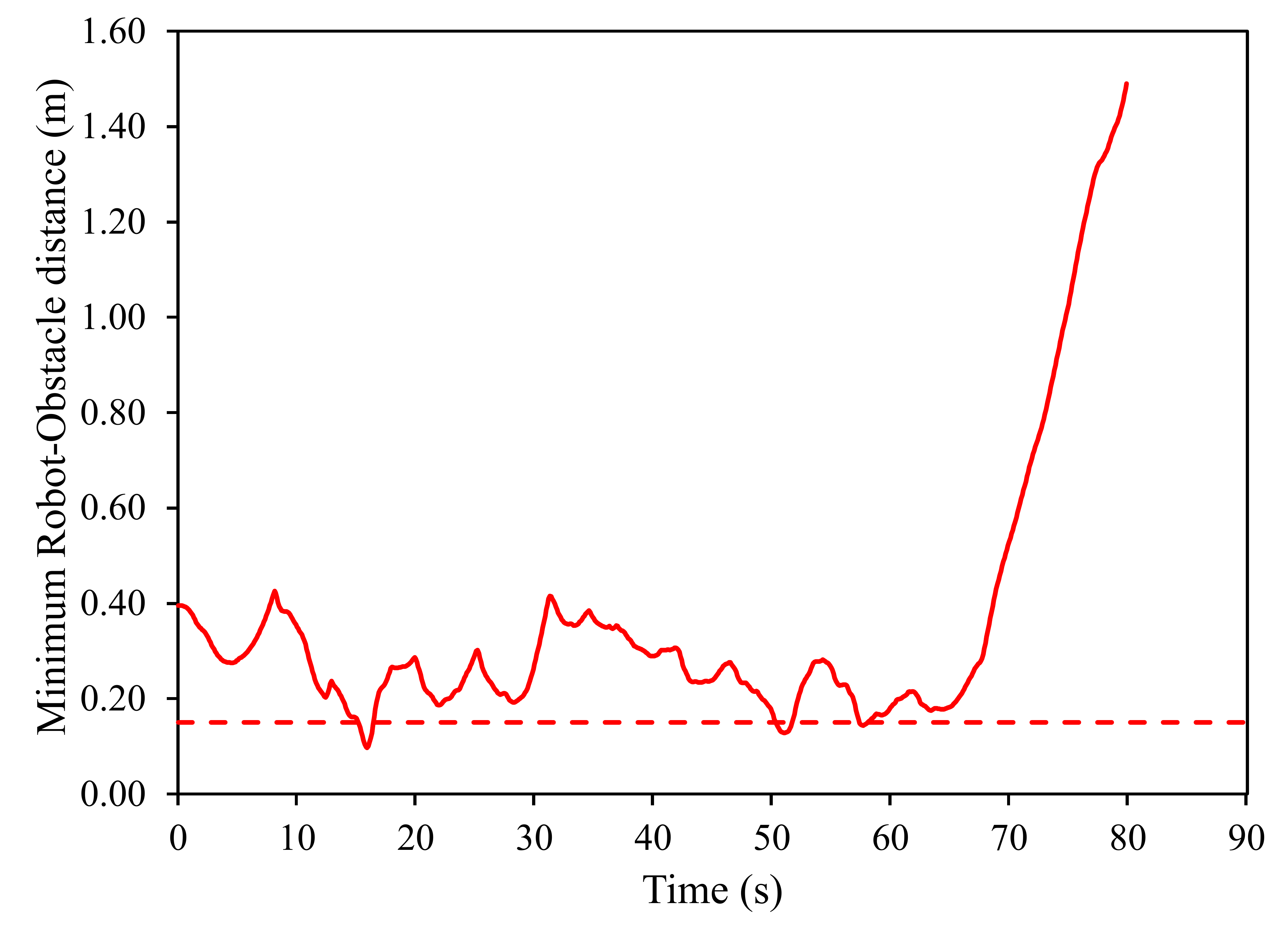}}
   \subfloat[]{\includegraphics[width=0.33\linewidth]{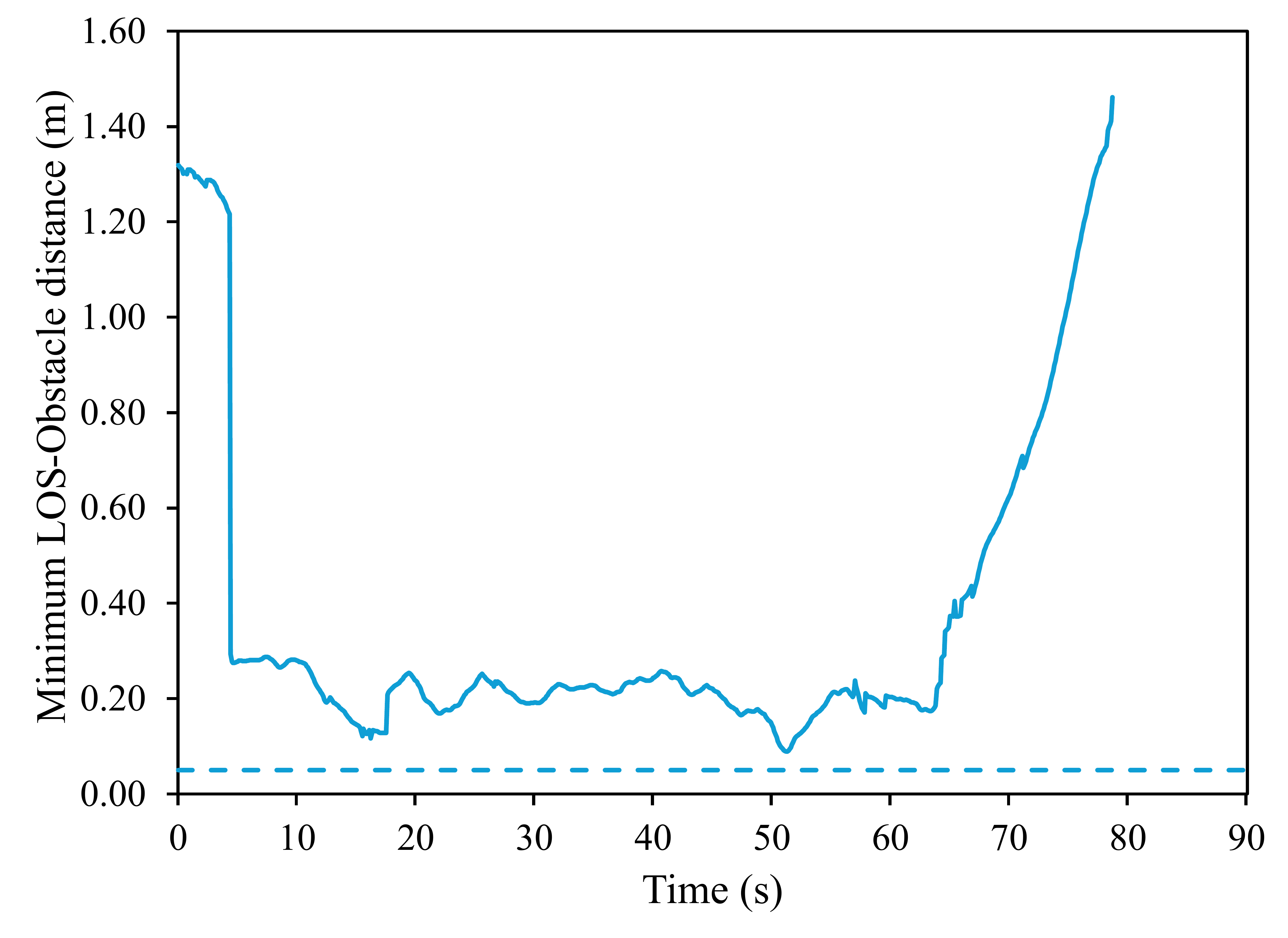}}
   \caption{Experiment results: (a) Minimum and maximum distance between the robots in link, (b) Minimum distance between the robot and obstacle, and (c) Minimum distance between the LOS between robots and obstacle at each time instances.
   The dashed lines in the figures represent the allowable minimum or maximum distances defined in (14)-(17).} 
   \label{fig:exp_results}
\end{figure*}

Figure \ref{fig:exp_snaps} presents snapshots of the experiment using the approximate method in Section \ref{sec:appr}.
Figure \ref{fig:exp_results} (a) and (b) illustrate the experiment's constraint enforcement status.
These results show that although the constraints given in \eqref{eq:c1}--\eqref{eq:c4} were violated a few times, the robot passed through the hole without collisions by combination with the recovery control in Section \ref{sec:recoverry}.

\section{Conclusion} \label{sec:concl}
In this study, we introduced a novel CBF-based control method for guiding a swarm of UAVs through obstacle-rich environments, while maintaining network connectivity without explicit communication between UAVs.
The proposed approach addresses the limitations of existing APF-based methods, such as oscillatory behaviors and frequent constraint violations, by utilizing CBF-based constraints instead of repulsive APFs and optimizing control inputs through a numerical optimization problem with soft constraints.
Additionally, an approximate method without numerical optimization was presented.
The effectiveness of these methods was validated through extensive simulations and real-world experiments using quadrotors, demonstrating improved performance in reducing vibratory movements compared to APF-based methods.
Both proposed approaches showed comparable performance; however, the approximate method was preferable to the optimization-based method since it required less computation time, making it advantageous for quadrotors with limited computational power.
Future work will explore the application of the proposed method in more complex environments involving larger swarms of UAVs equipped with onboard sensing and computation capabilities.

\section*{Funding}
This work is supported in part by JSPS KAKENHI grant number 22K04171

\bibliographystyle{tfnlm}
\bibliography{refs}

\end{document}